\documentclass{article} 
\usepackage{iclr2027_conference,times}
\usepackage{svg}

\usepackage{amsmath,amsfonts,bm}

\def\eqref#1{equation~\ref{#1}}

\def\1{\bm{1}}

\DeclareMathAlphabet{\mathsfit}{\encodingdefault}{\sfdefault}{m}{sl}
\SetMathAlphabet{\mathsfit}{bold}{\encodingdefault}{\sfdefault}{bx}{n}

\usepackage{wrapfig}
\usepackage{makecell}
\usepackage{diagbox}
\usepackage{array}
\usepackage{hyperref}
\usepackage{url}
\usepackage{graphicx}
\usepackage{booktabs}
\usepackage{multirow}
\usepackage{graphicx}
\usepackage{subcaption}

\usepackage{xcolor}
\usepackage{tcolorbox}
\usepackage{listings}
\tcbuselibrary{skins,breakable}
\definecolor{PromBoxGreen}{HTML}{65A64B}
\definecolor{PromBoxBackground}{HTML}{E8F3ED}
\definecolor{PromptBoxPurple}{HTML}{000086}
\definecolor{PromptBoxBackground}{HTML}{E6E6FF}
\newtcolorbox{promcopilotbox}[1]{%
  enhanced,breakable,
  colback=PromBoxBackground,colframe=PromBoxGreen,
  colbacktitle=PromBoxGreen,coltitle=white,
  fonttitle=\small\bfseries,fontupper=\small,
  title={#1},boxrule=0.7pt,arc=2mm,
  left=7pt,right=7pt,top=6pt,bottom=6pt,
  toptitle=4pt,bottomtitle=4pt,
  before skip=9pt,after skip=9pt,
  before upper={\setlength{\parindent}{0pt}\setlength{\parskip}{4pt}}
}

\newtcolorbox{deepeyebox}[1]{%
  enhanced,
  colback=PromBoxBackground,colframe=PromBoxGreen,
  colbacktitle=PromBoxGreen,coltitle=white,
  fonttitle=\small\bfseries,fontupper=\fontsize{9}{10}\selectfont,
  title={#1},boxrule=0.7pt,arc=2mm,
  left=7pt,right=7pt,top=6pt,bottom=6pt,
  toptitle=4pt,bottomtitle=4pt,
  before skip=9pt,after skip=9pt,
  before upper={\setlength{\parindent}{0pt}\setlength{\parskip}{1pt}}
}

\definecolor{DeepEyeFailureRed}{HTML}{CC0000}
\definecolor{DeepEyeFailureBackground}{HTML}{FDE7E7}
\newtcolorbox{deepeyefailurebox}[1]{%
  enhanced,
colback=DeepEyeFailureBackground,colframe=DeepEyeFailureRed,
  colbacktitle=DeepEyeFailureRed,coltitle=white,
  fonttitle=\small\bfseries,fontupper=\small,
  title={#1},boxrule=0.7pt,arc=2mm,
  left=7pt,right=7pt,top=6pt,bottom=6pt,
  toptitle=4pt,bottomtitle=4pt,
  before skip=9pt,after skip=9pt,
  before upper={\setlength{\parindent}{0pt}\setlength{\parskip}{3pt}}
}
\lstdefinestyle{deepeyequery}{%
  basicstyle=\ttfamily\footnotesize,
keywordstyle=\color{blue},stringstyle=\color{DeepEyeFailureRed},
  morekeywords={SELECT,DISTINCT,FROM,AS,WHERE,AND},
  morestring=[b]",morestring=[b]',
columns=fullflexible,keepspaces=true,breaklines=true,
showstringspaces=false,aboveskip=2pt,belowskip=3pt
}

\newtcolorbox{promptbox}[1]{%
  enhanced,breakable,
  colback=PromptBoxBackground,colframe=PromptBoxPurple,
  colbacktitle=PromptBoxPurple,coltitle=white,
  fonttitle=\small\bfseries,fontupper=\small,
  title={#1},boxrule=0.7pt,arc=2mm,
  left=7pt,right=7pt,top=6pt,bottom=6pt,
  toptitle=4pt,bottomtitle=4pt,
  before skip=9pt,after skip=9pt,
  before upper={\setlength{\parindent}{0pt}\setlength{\parskip}{4pt}}
}

\title{TQTS-Bench: A Multi-Syntax Benchmark for \\ Text-to-Query over Time-Series Databases}

\author{Fei Lyu\thanks{Equal contribution.},
  Zhiyi Peng\footnotemark[1], Jiaming Liu, Yixuan Yang, Changjian Chen\thanks{Corresponding authors.}, Zhuo Tang\footnotemark[2], \\ \textbf{Jiapeng Zhang, Kenli Li} \\
  College of Computer Science and Electronic Engineering, Hunan University  
}

\newcommand{\bench}{\textsc{TQTS-Bench}}
\iclrfinalcopy 
\begin{document}

\maketitle

\begin{abstract}
Large language models (LLMs) have significantly advanced natural language querying over relational databases, yet their ability to query time-series databases (TSDBs) remains largely unassessed. 
Existing benchmarks fail to adequately capture the non-unified query syntaxes, diverse application domains, and unique time-specific query intents inherent to TSDBs. 
To address this gap, we introduce \bench, a multi-syntax benchmark for evaluating text-to-query capabilities over TSDBs. 
\bench~contains 6,125 high-quality question--answering (QA) pairs spanning 97 TSDBs, 23 distinct query syntaxes, 22 application domains, and 4 types of time-specific query intents.
It is constructed through a human-centric AI-assisted workflow, where all QA pairs are carefully reviewed and revised by domain experts to ensure quality and correctness. 
Extensive evaluations of advanced LLMs and state-of-the-art text-to-query methods reveal challenges in querying TSDBs. 
Even the best-performing model evaluated, Claude-Opus-5, achieves only 48.98\% execution accuracy, while humans reach 87.34\%.
Error analysis reveals that this performance gap mainly stems from the heterogeneous query syntaxes across different TSDBs, misinterpretation of time-specific intents, and incorrect schema linking.
These findings highlight new opportunities to narrow the gap between current LLM capabilities and the requirements of TSDB queries in real-world applications.
The benchmark is available at: \url{https://anonymous.4open.science/r/TQTS-Bench-00CD}.
\end{abstract}

\section{Introduction}
Time-series data are widely used in real-world applications, such as the Internet of Things (IoT), AIOps, and cloud services. 
This widespread adoption has driven the rapid development of specialized time-series databases (TSDBs)~\citep{jensen2017time}.
However, analyzing data stored in TSDBs remains challenging because retrieving such data requires learning specialized query syntax~\citep{dranca2026llms}, particularly for users without expertise in databases.
Recently, due to the rapid advancement of large language models (LLMs), text-to-query techniques have progressed significantly, approaching human-level performance on relational database (RDB) benchmarks (e.g., Spider~\citep{yu2018spider} and BIRD~\citep{li2023bird}). 
This progress has substantially lowered the barrier to analyzing data stored in databases.
More recently, a few studies, such as~\citet{zhang2026promcopilot} and~\citet{dranca2026llms}, have begun to develop \textbf{T}ext-to-\textbf{Q}uery techniques specifically for \textbf{TS}DBs (\textbf{TQTS}).
However, it remains unclear how far these methods have advanced TQTS capabilities due to the lack of a comprehensive TQTS benchmark comparable to Spider and BIRD for RDBs.

\begin{figure}[t]
    \centering
    \includegraphics[width=1\linewidth]{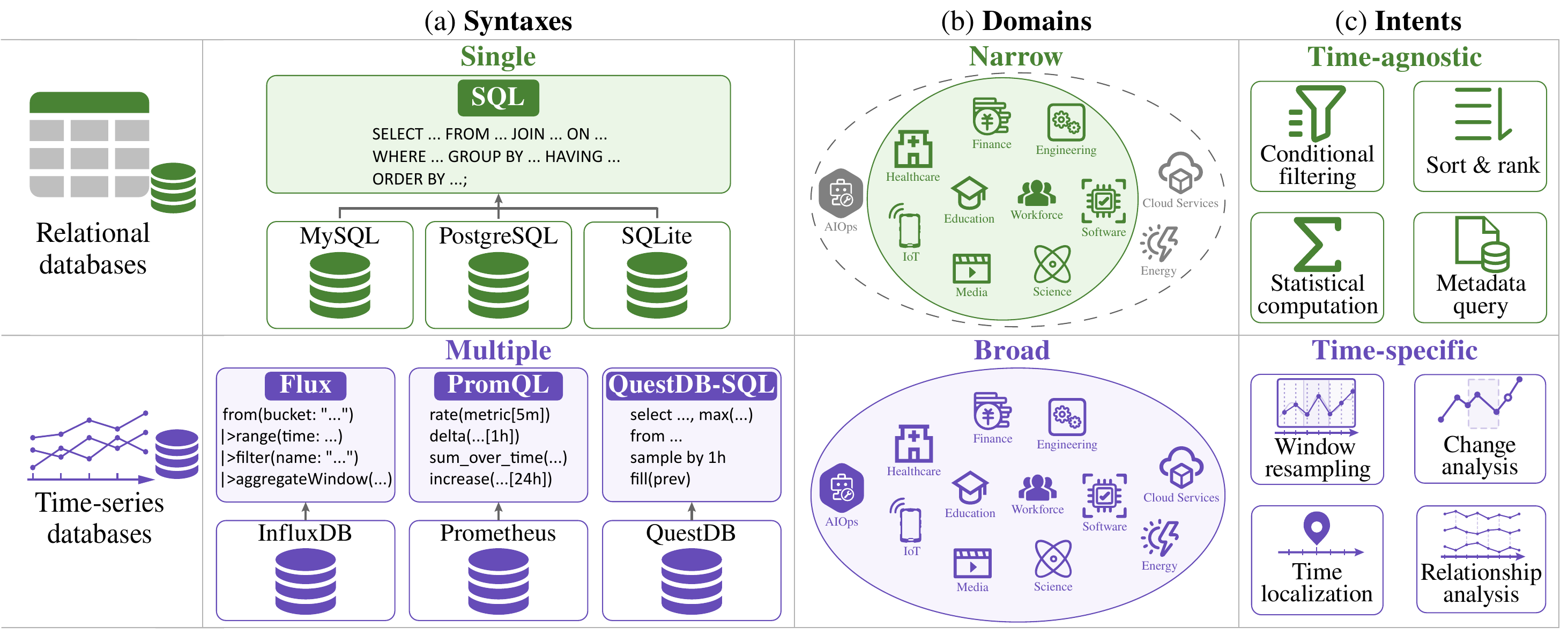}
    \caption{The three key differences between RDBs and TSDBs: (a) \textbf{Syntaxes}: unified SQL vs. diverse, non-uniform query syntaxes. (b) \textbf{Domains}: narrow domain coverage vs. broad domain coverage. (c) \textbf{Intents}: time-agnostic operations vs. time-specific analysis.}
    \label{fig:intro}
\end{figure}

Despite its importance, constructing a TQTS benchmark is not a straightforward extension of text-to-query benchmarks for RDBs, owing to three key differences (Fig.~\ref{fig:intro}).
\begin{itemize}
    \item \textbf{Syntaxes}: For RDBs, their query languages share the common SQL syntax with differences mostly limited to small dialects. However, for TSDBs, their query syntaxes can be substantially different (Fig.~\ref{fig:intro}(a)). For example, Flux in InfluxDB adopts a pipeline-style syntax, whereas PromQL in Prometheus relies on nested, function-oriented expressions. Moreover, even TSDBs that stay close to SQL, such as QuestDB, extend it with a wide range of time-specific operations and functions, further diverging from basic SQL syntax.
    \item \textbf{Domains}: In domains that require the continuous collection of high-frequency data, such as IoT and cloud services, TSDBs are predominant. Because RDBs are not specifically designed for such workloads, existing text-to-query benchmarks for RDBs rarely cover these domains, resulting in limited domain coverage (Fig.~\ref{fig:intro}(b)).
    \item \textbf{Intents}: Queries over RDBs typically focus on time-independent intents such as filtering, sorting, and aggregation. In contrast, TSDB queries involve many time-specific intents, such as change analysis and relationship analysis (Fig.~\ref{fig:intro}(c)). This makes text-to-query for RDBs and TSDBs differ in both question and intent understanding.
\end{itemize}

Due to the three differences above, rather than simply extending existing text-to-query benchmarks for RDBs, we construct \textsc{TQTS-Bench}, a new benchmark specifically designed for text-to-query over TSDBs.
\textsc{TQTS-Bench} contains 6,125 text-to-query question-answering (QA) pairs, spanning 97 TSDBs with \textbf{23 distinct query syntaxes}.
These TSDBs span \textbf{22 application domains}, while the QA pairs cover \textbf{4 time-specific query intents}.
To construct this benchmark, we first identified 23 popular and representative TSDB management systems based on the well-known \textit{DB-Engines Ranking of TSDB Management System}\footnote{\url{https://db-engines.com/en/ranking/time+series+dbms}} 
and collected 427 datasets and 1,405 seed QAs related to these TSDB management systems from public sources such as TSDB official documentation, forums, and technical posts.
These 427 datasets were manually filtered based on data authenticity and schema complexity, resulting in 97 datasets.
In collaboration with domain experts, we analyzed the 1,405 seed QAs and identified 9 representative query intents.
Based on these query intents and the seed QAs, we developed a human-centric AI-assisted workflow supported by a visual analytics tool to construct new QA pairs. 
Finally, these QA pairs underwent a rigorous cross-validation process to ensure their correctness and quality, and 6,125 QA pairs were verified.


We conduct a comprehensive evaluation of advanced LLMs and representative text-to-query methods on \bench. 
The results show that current LLMs still struggle with TQTS tasks.
For example, the best-performing model, Claude-Opus-5, achieves an execution accuracy of only 48.98\%, which remains far from human performance (87.34\%), highlighting the challenges posed by this benchmark.
Meanwhile, both existing RDB methods and TSDB methods also achieve limited performance, suggesting that existing methods still face difficulties in TQTS tasks and exhibit limited transferability and generalizability across different settings.
We further conduct an error analysis and identify three main factors behind these failures, including heterogeneous query syntaxes across different TSDBs, misinterpretation of time-specific intents, and incorrect schema linking. 
Based on these findings, we discuss opportunities and promising directions for improving TQTS performance to enable more effective time-series data analysis in real-world applications.

\section{Related Work}
\label{related}

\paragraph{\textbf{Text-to-query benchmarks over RDBs}.}    
RDB benchmarks have established well-defined evaluation paradigms for the Text-to-SQL task. 
Spider~\citep{yu2018spider} evaluates whether models can generalize to unseen databases in different domains. 
BIRD~\citep{li2023bird} extends this setting to larger and real-world databases, requiring models to use database contents and external knowledge. 
Spider~2.0~\citep{lei2025spider2} further targets complex enterprise environments with large schemas and realistic query workflows.
Beyond these general-purpose benchmarks, several studies focus on specific applications.
For example, ScienceBenchmark~\citep{zhang2023sciencebenchmark} targets scientific databases, while SParC~\citep{yu2019sparc} and CoSQL~\citep{yu2019cosql} focus on contextual and conversational queries.
Additionally, many benchmarks~\citep{vo2022temporal,chang2023drspider,liu2026logiccat} evaluate model capabilities on reasoning and query robustness.
Despite covering broad evaluation settings, these benchmarks are grounded in RDBs and ignore time-series query requirements.
However, TSDBs inherently contain multiple query syntaxes and are designed for time-series analysis. 
Therefore, existing RDB benchmarks are insufficient for evaluating text-to-query over TSDBs.

\paragraph{\textbf{Text-to-query benchmarks over TSDBs}.}
Recent studies have developed several text-to-query benchmarks for specific TSDBs.
For example, TEFD~\citep{wu2026tefd} evaluates text-to-Flux generation for InfluxDB, while another study~\citep{dranca2026llms} evaluates the generation of both Flux and InfluxQL queries in an agro-food production scenario.
PromCopilot~\citep{zhang2026promcopilot} focuses on generating PromQL queries for Prometheus-based system monitoring. 
However, these studies are often limited to specific TSDBs, query syntaxes, or application domains, preventing a comprehensive evaluation of LLMs' text-to-query capabilities across diverse TSDB environments and scenarios. 
To address this gap, we develop \bench, a multi-syntax benchmark that spans a broad range of TSDBs, query syntaxes, and application domains, enabling a systematic and comprehensive evaluation of LLMs for text-to-query tasks over TSDBs.
\section{Dataset Construction}
\label{dataset}

\subsection{Task Definition}
Given a natural-language question, a database schema, and a target TSDB, the TQTS task aims to generate an executable query that correctly retrieves the requested data from the TSDB.
Here, the database schema describes the logical organization of time-series data, including how timestamps, data values, and associated metadata are structured and typed~\citep{bader2017survey}. 
Depending on the TSDB management system, these elements may be represented as tables or measurements, columns or fields, and tags or labels.
The target TSDB refers to a dataset stored in a specified TSDB management system (e.g., InfluxDB).

\subsection{Construction pipeline}
\begin{figure}[htbp]
    \centering
    \includegraphics[width=1\linewidth]{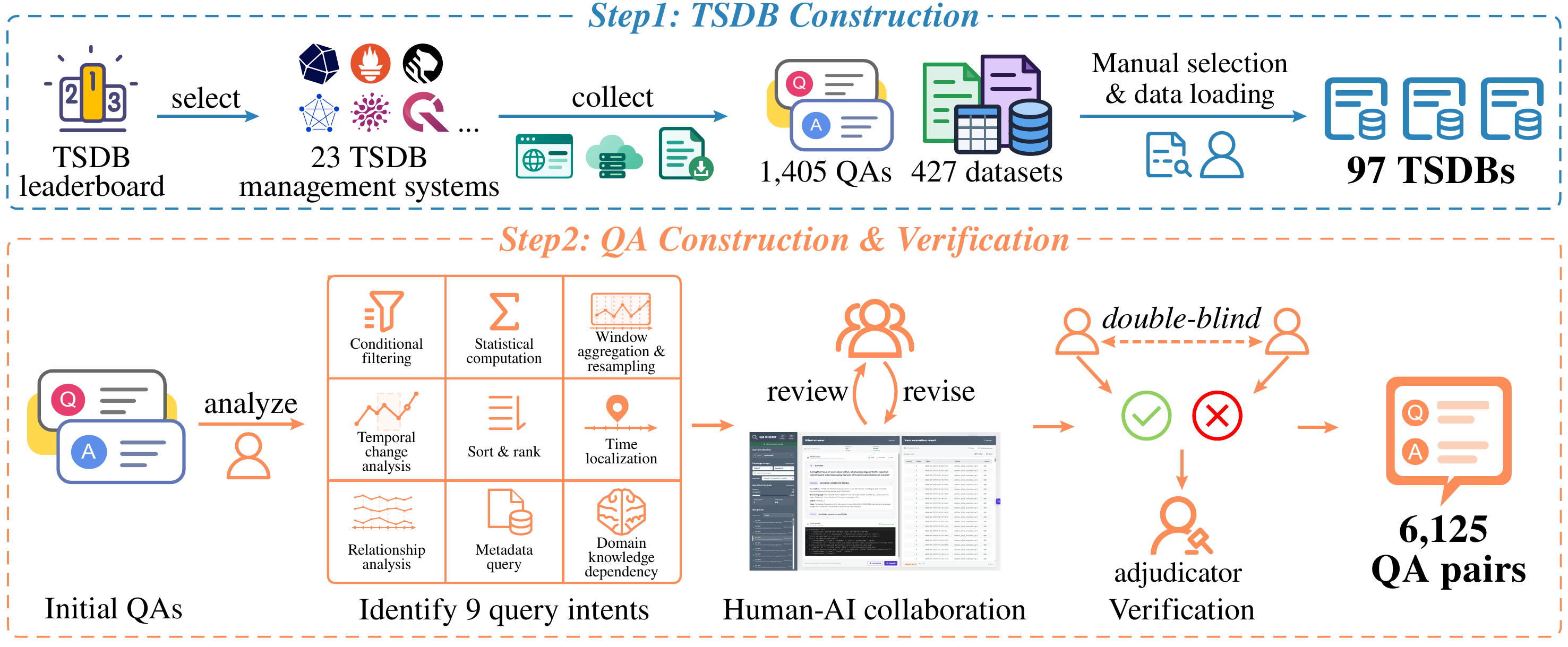}
    \caption{The construction pipeline of \bench.}
    \label{fig:pipeline}
\end{figure}

As shown in Fig.~\ref{fig:pipeline}, the construction of \bench~consists of two main steps: TSDB construction and QA construction and verification.
The TSDB construction step collects and curates 97 high-quality TSDBs, and the QA construction and verification step constructs 6,125 QA pairs with a human-centric AI-assisted workflow.

\subsubsection{TSDB construction}
Based on the well-known \textit{DB-Engines Ranking of TSDB Management System}, we first select 23 popular and representative TSDB management systems depending on the accessibility.
With these systems, we then collect real-world TSDB datasets from diverse application scenarios.
Although existing studies~\citep{li2023bird,yu2018spider,zhang2026promcopilot,wu2026tefd} provide datasets for text-to-query tasks, these datasets are typically distributed as standalone files (e.g., CSV files) rather than as operational TSDBs.
Therefore, we gather publicly available datasets from multiple real-world sources, including official websites, hosting platforms that export data from specific TSDBs, and datasets documented as being collected or stored using TSDBs.
In parallel, we compiled a set of seed QAs to facilitate subsequent identification of the query intent. 
These QA pairs were sourced from official TSDB documentation, community forum discussions, and technical posts.

The collection process results in 427 candidate TSDB datasets and 1,405 seed QAs.
We then manually inspect these candidate datasets based on data authenticity and schema complexity, and finally select 97 representative source datasets.
Subsequently, we import these datasets into their corresponding TSDB management systems to instantiate the database environments used in our benchmark.
For TSDB details, refer to App.~\ref{appendix:tsdb-information}.

\subsubsection{QA construction and verification}
In \bench, each QA pair consists of a natural-language question and its corresponding answer for a specific TSDB. 
To construct the QA pairs, we first identify representative query intents from the seed QAs. 
We also developed a visual analytics tool that supports a human-centric AI-assisted workflow to help annotators directly explore and interact with the underlying TSDB.

\textbf{Query intent identification}.
With the collected seed QAs, we collaborate with domain experts to analyze the underlying query requirements in these pairs and common TSDB usage scenarios.
Based on this analysis, we identify nine representative query intents, comprising four time-specific query intents (\textbf{I1}--\textbf{I4}) and five time-agnostic query intents (\textbf{I5}--\textbf{I9}). 
Specifically, the time-specific intents are: \textbf{I1}: Window aggregation \& resampling; \textbf{I2}: Temporal change analysis; \textbf{I3}: Time localization; and \textbf{I4}: Relationship analysis. 
The time-agnostic intents are: \textbf{I5}: Conditional filtering; \textbf{I6}: Statistical computation; \textbf{I7}: Sort \& rank; \textbf{I8}: Metadata query; and \textbf{I9}: Domain knowledge dependency.
We find that most query requirements can be represented by a single intent or a combination of multiple intents, indicating that the identified intents cover a broad range of TSDB usage needs.
These nine query intents provide a common foundation for QA construction across different TSDB systems, regardless of their underlying query syntax.
For details, refer to App.~\ref{appendix:query-intent}.

\textbf{Human-AI collaborative construction}.
In \bench, a qualified QA pair should consist of a natural-language question that faithfully reflects query intents in the target TSDB scenario, together with an executable query that correctly answers it under the given schema and TSDB context.
To facilitate QA construction, we develop a visual analytics tool that supports a human-centric AI-assisted workflow.
Specifically, we provide LLMs (e.g., GPT-5.6-Sol) with the identified query intents, database schemas, and TSDB context, requiring them to generate the corresponding questions.
Annotators then review these questions to verify they match the query intents and reflect meaningful TSDB scenarios, refining their descriptions when necessary to improve clarity and alignment with the query intents.
If the questions are determined, the annotators then use LLMs to generate the corresponding executable queries and construct the generated unverified QA pairs.

\textbf{QA verification}.
After construction, all generated unverified QA pairs undergo a final verification step to ensure their correctness and quality. 
To facilitate this process, we use the developed visual analytics tool.
For each unverified QA pair, we conduct a cross-validation procedure.
Specifically, two annotators independently hand-write queries based on the question, schema, and provided TSDB context.
They can execute the hand-written queries to determine whether the query results match the question's requirements.
If both annotators obtain results consistent with the unverified answer and both determine that the question is answerable on the target TSDB, the QA pair is accepted as verified.
Otherwise, the case is escalated to an adjudicator. 
The adjudicator examines the annotators' queries, results, and feedback, and revises the QA pairs accordingly. 
The revised QA pairs then undergo a further round of verification, or are finalized by the adjudicator if the issue is resolved.
For the visual analytics tool and verification details, refer to App.~\ref{appendix:interface}.

Overall, a total of 6,125 QA pairs are verified. 
During this process, the annotators collectively spent more than 600 hours and executed 30,774 queries, averaging approximately 6 minutes inspection and 5 executions per QA pair. 
This rigorous verification process facilitates careful inspection and iterative validation, thereby ensuring the quality and correctness of the constructed QA pairs.

\subsection{Data Statistics}
\bench~is a multi-syntax benchmark designed for TQTS tasks.
Specifically, we summarize the data statistics of \bench~from three perspectives: supported query syntaxes, domain coverage, and time-specific query intents.

\begin{figure}[htbp]
    \centering
    \includegraphics[width=1\textwidth]{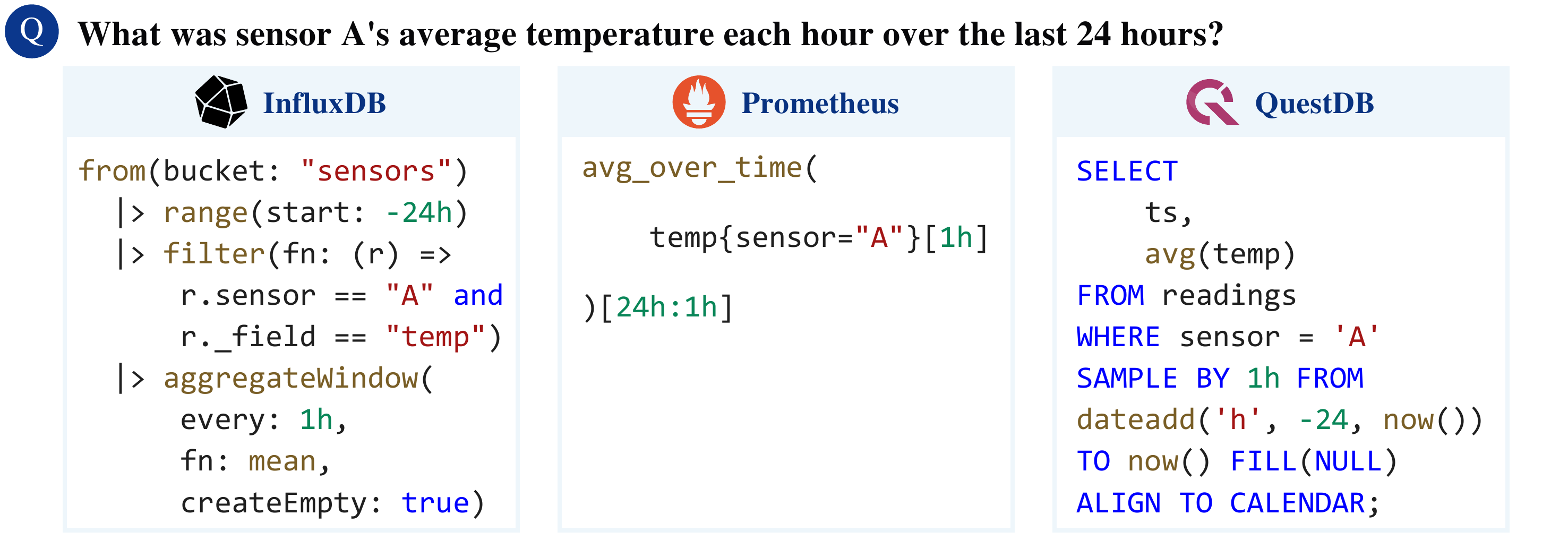}
    \caption{A representative question expressed using Flux, PromQL, and QuestDB SQL.}
    \label{fig:multi-syntax}
\end{figure}

\textbf{Multiple query syntaxes}.
In total, \bench~contains 6,125 QA pairs spanning 23 distinct query syntaxes across 97 TSDBs, which differ substantially in their structures and usage patterns. 
As shown in Fig.~\ref{fig:multi-syntax}, for the same question, InfluxDB's Flux represents the query as a pipeline of transformations, Prometheus's PromQL uses metric selectors and range-vector functions, and QuestDB SQL extends conventional SQL with dedicated temporal operations such as time sampling. 
This diversity makes \bench~challenging, as models must generate queries that are both semantically correct and syntactically compatible with the target TSDB.
For more details, refer to App.~\ref{appendix:tsdb-ms-and-query-syntax}.

\begin{figure}[htbp]
  \centering
  \includegraphics[width=1\textwidth]
      {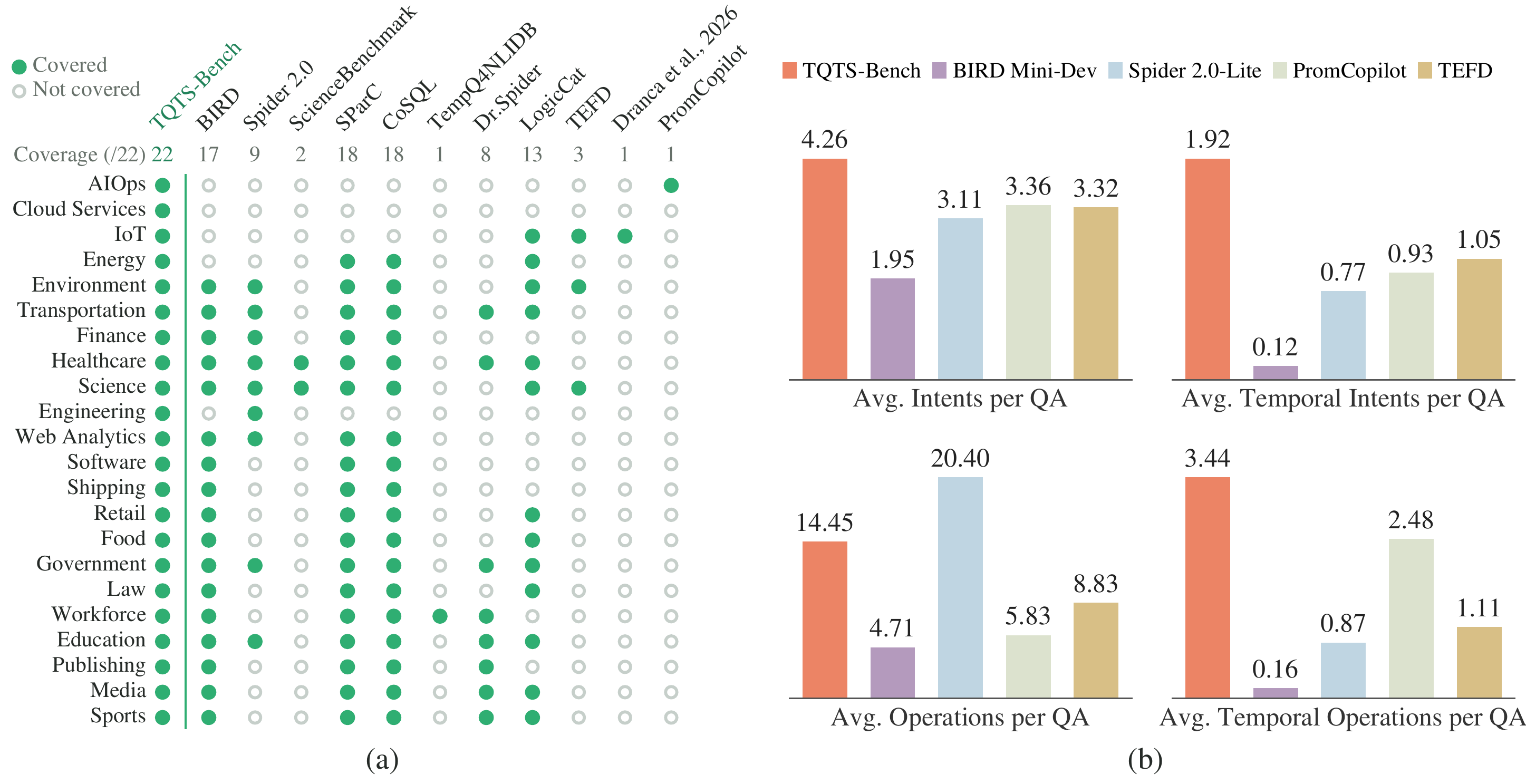}
  \caption{
(a) Domain coverage and (b) time-specific query intent statistics of \bench.}
  \label{fig:domain_intent_statistics}
\end{figure}
\textbf{Broad domain coverage}.
Fig.~\ref{fig:domain_intent_statistics}(a) compares the domain coverage of \bench~with existing text-to-query benchmarks over RDBs and TSDBs.
\bench~covers all 22 application domains considered in the comparison, while previous benchmarks typically cover only a subset of them.
In particular, many RDB benchmarks emphasize general domains such as finance, healthcare and education. 
In contrast, TSDB benchmarks are often concentrated on scenarios involving continuously generated, high-frequency time-series data, such as IoT, AIOps, and cloud services, where efficient ingestion and temporal analysis are critical.
By spanning both general-purpose and time-intensive domains, \bench~supports a more comprehensive evaluation of text-to-query methods across real-world data settings.
For more domain details, refer to App.~\ref{appendix:domain}

\textbf{More time-specific query intents}.
Fig.~\ref{fig:domain_intent_statistics}(b) reports the query intent statistics across different text-to-query benchmarks. 
Compared with other benchmarks, \bench~has the highest average numbers of both query intents and time-specific intents per QA pair, as well as the highest average number of time-specific operations per QA pair. 
Together, these statistics highlight the temporal complexity of \bench, making it a challenging benchmark for translating diverse time-specific intents into executable queries.
For more query intent details, refer to App.~\ref{appendix:query-intent}.

\section{Experiment}
\label{experiment}
\subsection{Experimental Setup}

\textbf{Evaluation metric}.
We evaluate model performance using the widely adopted metric \textbf{Execution Accuracy (EX)}~\citep{li2023bird,lei2025spider2}, which measures whether the execution result matches the expected answer and thereby reflects the semantic correctness of the generated query.

\textbf{Difficulty level}.
Given the diversity of query syntaxes, token count of the gold query for a question is not a suitable measure of difficulty. 
Therefore, we classify questions by the number of query intents: 1--3 as easy (31.72\%), 4--5 as medium (44.33\%), and 6--9 as hard (23.95\%).
This categorization reflects the increasing compositional complexity required to answer the question.

\textbf{Human evaluation}. 
Given the substantial time and cost involved, we conduct human evaluation on a randomly sampled 10\% subset of \bench~to establish a human performance baseline for real-world TSDB application scenarios.
For details, refer to App.~\ref{appendix:human-evaluation}.

\textbf{LLMs}.
We evaluated a broad range of advanced LLMs, including both open-source and closed-source models. 
These models comprise representative and state-of-the-art offerings from major AI developers, enabling a comprehensive comparison between open-source and proprietary frontier models.
The open-source models include Qwen3.8-Flash~\citep{qiu2026designqwen38nextarchitectureevaluation}, DeepSeek-V4-Pro~\citep{deepseekv4}, GLM-5.3-Flash~\citep{glm5}, and Kimi-K3~\citep{kimik3}, while the closed-source models include GPT-6-Sol~\citep{openai2026gpt6sol}, Claude-Opus-5~\citep{anthropic2026claudeopus5}, and Gemini-3.7-Flash~\citep{gemini37flash}.
Following the experiment setting of~\citet{lei2025spider2}, if the input length exceeds the model's maximum token limit, it will be truncated from the beginning.

\textbf{Text-to-query methods}.
We also evaluate current state-of-the-art and representative text-to-query methods for both RDBs and TSDBs to examine their ability to address TSDB-specific challenges and transferability across different TSDB management systems. 
The RDB methods include DeepEye-SQL \citep{li2026deepeyesql},  OpenSearch-SQL \citep{xie2025opensearchsql}, RSL-SQL \citep{cao2024rslsql}, DAIL-SQL \citep{gao2024dailsql}, and DIN-SQL \citep{pourreza2023dinsql}, while the TSDB methods include PromCopilot \citep{zhang2026promcopilot}.
Although RDB methods are not specifically designed for TSDBs, they provide strong baselines because they address fundamental text-to-query challenges such as schema linking and query generation, which are also critical in TSDB applications. 
To ensure a controlled comparison and eliminate the impact of different backbone models, we use GPT-4o-mini as the unified LLM backbone for all methods.


\subsection{Quantitative Evaluation Results}
\begin{table}[htbp]
\centering
\caption{Evaluation results on \bench. The best result (except human) is in \textbf{bold}, and the runner-up is \underline{underlined}. Results with (*) are tested on a randomly sampled 10\% subset.}
\label{tab:llm-results}
\begin{tabular}{lcccc}
\hline
\multirow{2}{*}{\textbf{Method}} & \multicolumn{4}{c}{\textbf{EX($\uparrow$)}} \\ \cline{2-5} 
                                 & Easy  & Medium  & Hard  & \textbf{Overall}  \\ \hline
Human performance                & *95.71\%  & *85.52\%     & *81.94\%  & *87.34\%                 \\ \hline
\multicolumn{5}{c}{\textit{\textbf{Open-Source Models}}}                                     \\ \hline
Qwen3.8-Flash                    & 43.75\%    & 20.07\%        & 17.79\%      & 27.04\%                \\
DeepSeek-V4-Pro                  & 39.89\%    & 15.87\%      & 11.79\%    & 22.51\% \\
GLM-5.3-Flash    & 54.30\%    & 28.88\%      & 19.15\%    & 34.61\%               \\
Kimi-K3                          & 58.26\%    & 37.24\%     & 34.08\%    & 43.15\%               \\ \hline
\multicolumn{5}{c}{\textit{\textbf{Closed-Source Models}}}                                     \\ \hline
GPT-6-Sol                      & 62.74\%    & \underline{40.63\%}      & \underline{35.38\%}    & \underline{46.38\%}               \\
Claude-Opus-5   & \underline{62.79\%}    & \textbf{42.91\%}     & \textbf{41.92\%}    & \textbf{48.98\%}\\
Gemini-3.7-Flash  & \textbf{63.25\%}    & 37.68\%      & 32.92\%    & 44.65\% 
               \\ \hline
\multicolumn{5}{c}{\textit{\textbf{RDB Methods}}}                              \\ \hline
DeepEye-SQL                      & 14.82\%    & 1.92\%      & 1.16\%    & 5.83\%                \\
OpenSearch-SQL                   & 16.37\%    & 1.84\%      & 1.09\%    & 6.27\%                \\
RSL-SQL                          & 11.99\%    & 1.33\%      & 0.95\%    & 4.62\%               \\
DAIL-SQL    & 1.34\%    & 0.26\%      & 0.00\%    & 0.54\%                \\
DIN-SQL                          & 12.25\%   & 2.17\%     & 1.23\%    & 5.14\%
               \\ \hline
\multicolumn{5}{c}{\textit{\textbf{TSDB Methods}}}                             \\ \hline
PromCopilot                      & 1.80\%    & 0.22\%      & 0.07\%    & 0.69\%               \\ \hline
\end{tabular}
\end{table}

\textbf{Existing RDB methods struggle with TQTS tasks and exhibit limited transferability to TSDBs.}
Tab.~\ref{tab:llm-results} reports the performance of state-of-the-art RDB methods on TQTS tasks.
Specifically, DeepEye-SQL, a high-performing RDB method among those compared on the BIRD benchmark\footnote{\url{https://bird-bench.github.io/}}, achieves only 5.83\% EX on \bench, a notably low score.
Moreover, other RDB methods also perform poorly, with most achieving below 10.00\% EX and some as low as 0.54\%.
This performance gap between RDB benchmark and \bench~suggests that methods optimized for RDB text-to-query tasks do not necessarily generalize to TSDB scenarios.
More detailed analysis of these results is provided in App.~\ref{appendix:rdb-methods}.

\textbf{Although LLMs outperform existing RDB methods, their performance remains unsatisfactory.}
Table~\ref{tab:llm-results} shows that advanced LLMs still achieve limited EX on TQTS tasks.
The best-performing model, Claude-Opus-5 achieves the highest EX and consistently outperforms other models at the medium and hard difficulty levels.
However, its overall performance remains relatively low, with an EX of only 48.98\%, which is far below the human performance of 87.34\%.
This gap indicates that TQTS tasks remain challenging even for the most advanced models.
The gap is more pronounced for some open-source models; for example, DeepSeek-V4-Pro achieves an EX of only 22.51\%, leaving substantial room for improvement in practical applications.
Compared with explicitly optimized RDB methods for text-to-SQL tasks, LLMs demonstrate stronger generalization ability, but their performance remains far from human-level capability.

%
\begin{wraptable}{r}{0.35\columnwidth}
\centering
\caption{Performance of PromCopilot on different domains.}
\label{tab:promcopilot}
\small
\setlength{\tabcolsep}{2pt}
\begin{tabular}{
    >{\raggedright\arraybackslash}p{1.8cm}
    >{\centering\arraybackslash}p{1.1cm}}
    \toprule
    Domain& EX($\uparrow$) \\
    \midrule
     Targeted& 3.03\% \\
     Others& 0.00\%\\
\bottomrule
\end{tabular}
\end{wraptable}
\textbf{Existing TSDB methods exhibit limited generalizability for TQTS tasks}.
Tab.~\ref{tab:llm-results} shows that PromCopilot, a state-of-the-art TSDB method for TQTS tasks, still achieves limited performance, with an EX of only 0.69\%.
PromCopilot is specifically designed for PromQL, the query syntax of Prometheus, and is primarily targeted in cloud services and AIOps application domains.
As reported in Tab.~\ref{tab:promcopilot}, its performance decreases from 3.03\% to 0.00\% when the domain shifts to others.
The result indicates that existing TSDB methods are sensitive to changes in application domains, revealing the limited generalizability of existing TSDB methods across different TQTS task settings. 
For detailed analysis, refer to App.~\ref{appendix:tsdb-methods}.


\subsection{Error analysis}
\label{subsec:error-analysis}
\begin{wrapfigure}[17]{r}{0.35\linewidth}
  \centering
  \includegraphics[width=\linewidth]{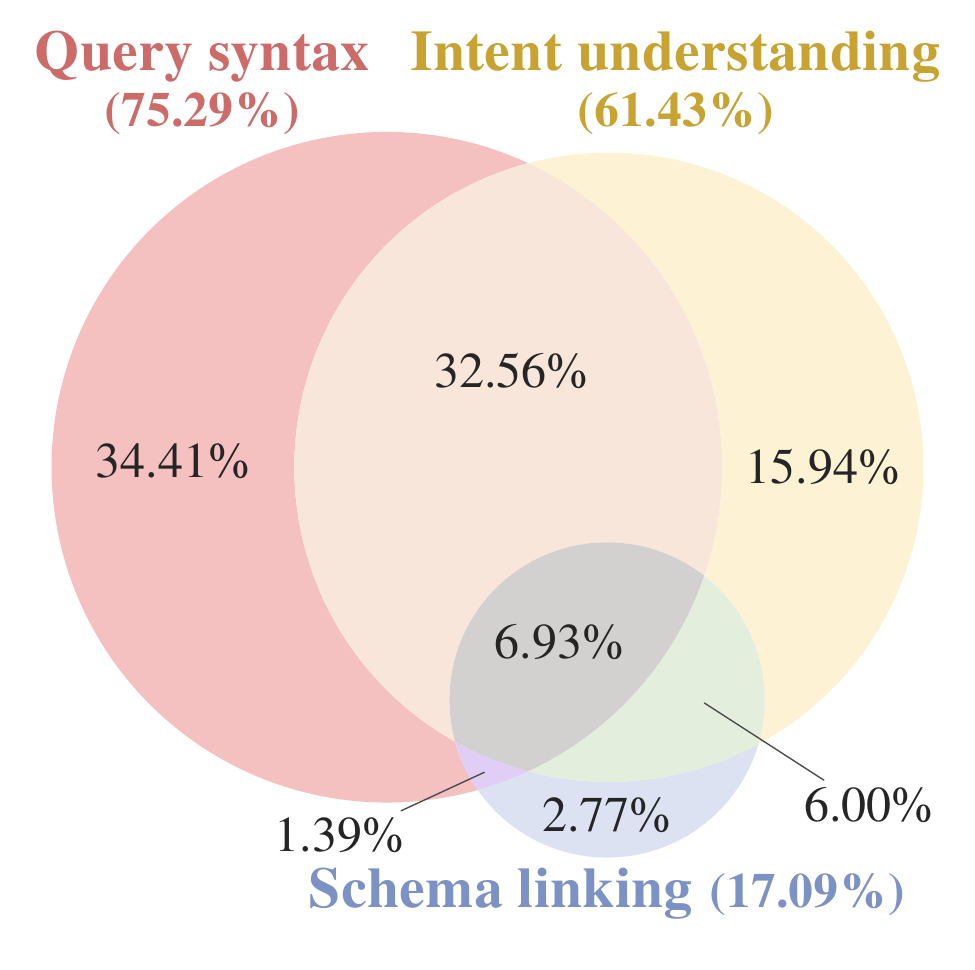}
  \caption{Proportions of incorrect cases by error type.}
  \label{fig:error-venn}
\end{wrapfigure}
We perform a systematic error analysis  by randomly sampling 600 examples and manually examining the incorrect cases.  
Given its weak performance, we select DeepSeek-V4-Pro for error analysis, as it may expose more representative failure modes in practical settings.
Fig.~\ref{fig:error-venn} shows three identified representative error types: \textbf{query syntax errors}, \textbf{intent understanding errors}, and \textbf{schema linking errors}.

\begin{figure}[htbp]
    \centering
    \includegraphics[width=\linewidth]{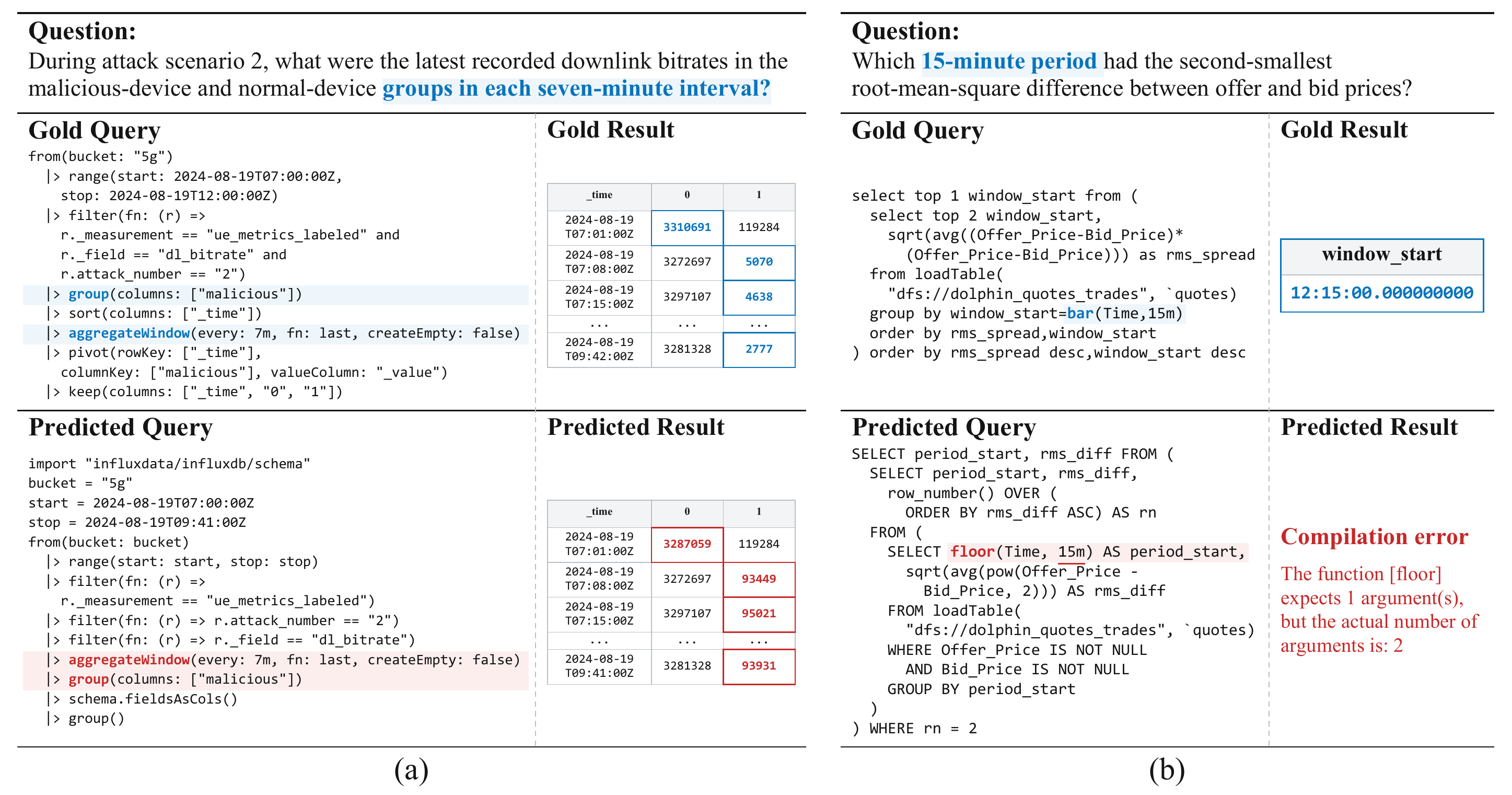}
    \caption{Examples of \textbf{query syntax errors}.
    (a) Query structure error: reversing \texttt{group} and \texttt{aggregateWindow} selects per-series rather than per-group last readings, producing incorrect results.
    (b) Function usage error: replacing \texttt{bar(Time, 15m)} with \texttt{floor(Time, 15m)} violates the single-argument requirement of \texttt{floor}, causing a compilation error.}
    \label{fig:error_syntax_case_study}
\end{figure}

\textbf{Query syntax errors (75.29\%).}
Due to the diversity and complexity of TSDB query syntax, the model needs to understand various syntax structural patterns and function invocation rules to generate correct queries.
We categorize query syntax errors into two types based on the type of syntax violation:
(1) \textbf{Query structure errors (42.50\%).}
The model fails to construct valid query structures such as improper ordering of operators or functions, resulting in incorrect query results (e.g., Fig.~\ref{fig:error_syntax_case_study}(a)).
(2) \textbf{Function/keyword usage errors (32.79\%).}
The model misuses functions or keywords, including incorrectly applying valid functions (i.e., misunderstanding parameters, or usage constraints) and invoking non-existent functions, thereby causing errors (e.g., Fig.~\ref{fig:error_syntax_case_study}(b)).
For more error cases, refer to App.~\ref{appendix:error-categories-examples}.

\begin{figure}
    \centering
    \includegraphics[width=1\linewidth]{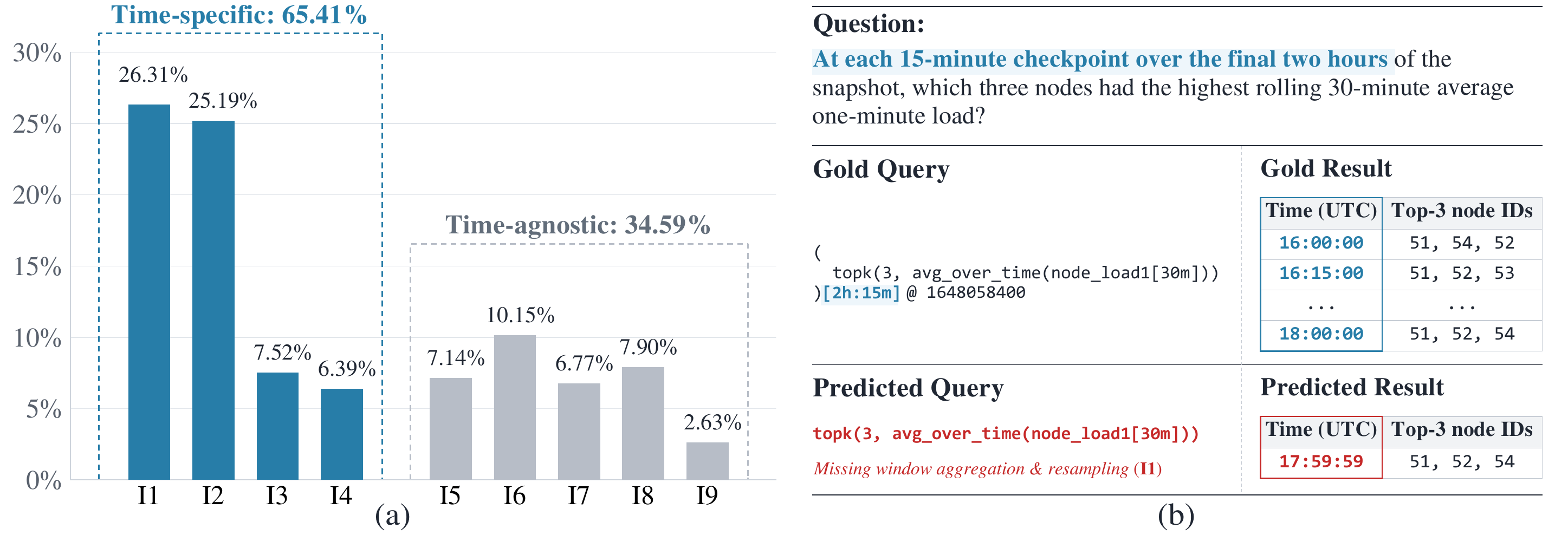}
    \caption{\textbf{Intent understanding errors}. (a) Distribution of primary erroneous intents; time-specific intents account for 65.41\%. (b) Window aggregation \& resampling error: the predicted query lacks \texttt{[2h:15m]}, returning a single top-3 result instead of results at each 15-minute checkpoint.}
    \label{fig:intent_errors_combined}
\end{figure}

\textbf{Intent understanding errors (61.43\%)}.
For the TQTS task, accurately identifying the requirement of a question is particularly challenging due to the complexity of time-specific query intents.
\bench~includes 9 query intents, 4 of which are time-specific and require models to correctly interpret and execute temporal operations.
As shown in Fig.~\ref{fig:intent_errors_combined}(a), errors involving time-specific query intents account for 65.41\% of all intent understanding errors, substantially exceeding those associated with time-agnostic query intents (34.59\%).
Specifically, the error rates for the 4 time-specific intents are 26.31\%, 25.19\%, 7.52\%, and 6.39\%, respectively. 
These results suggest that current models struggle particularly with interpreting time-specific query intents. 
Fig.~\ref{fig:intent_errors_combined}(b) shows one such failure, caused by missing window aggregation and resampling.

\textbf{Schema linking errors (17.09\%)}.
Schema linking is a critical step in text-to-query tasks~\citep{li2026deepeyesql}.
However, existing methods are mainly designed for RDBs and cannot be directly applied to TSDBs due to their different schema organizations.
TSDBs typically organize data using metrics, labels, and tags rather than tables and columns, making traditional schema linking error types unsuitable for TSDBs.
Moreover, TSDBs are widely used in IoT and AIOps domains, where high-frequency data collection leads to large-scale schemas.
We identify two major challenges for schema linking in TQTS tasks:
\textbf{(1) Diverse schema structures.}
The heterogeneous use of metrics, labels, and tags introduces new schema ambiguities. 
For schema details, refer to App.~\ref{appendix:diverse-schema}.
\textbf{(2) Large-scale schema.}
Cases with schema linking errors have much larger schemas (\textbf{381k tokens on average}) than other erroneous cases (115k tokens on average).
Large schemas hinder schema comprehension and relevant element retrieval, and long contexts can further degrade LLM performance due to the ``lost-in-the-middle'' phenomenon~\citep{liu2024lost}.
Therefore, improving schema linking performance in TQTS tasks remains an important direction for future work.

\subsection{Ablation study}
\label{subsec:ablation-study}
\begin{wraptable}[11]{r}{0.35\linewidth}
  \centering
  \small
  \caption{EX across different query syntaxes and benchmarks.}
  \label{tab:syntax-ablation}
  \begin{tabular}{p{0.37\linewidth} p{0.14\linewidth}
    p{0.14\linewidth}}
    \toprule
    Setting & Syntax & EX($\uparrow$) \\
    \midrule
    Original        & Diverse & 0.00\%     \\
    w/ conversion   & SQL      & 32.21\% \\
    BIRD            & SQL      & 56.91\%     \\
    {Spider~2.0-lite} & SQL      & 22.94\% \\
    \bottomrule
  \end{tabular}
\end{wraptable}
To figure out why existing models perform poorly in \bench, we conduct an ablation study. 
Although domains and query intents may also affect TSDB query generation, we focus on syntax because it is an intuitive and explicit difference between TSDBs and RDBs.
Specifically, we convert the original TSDBs into RDBs (e.g., SQLite) while keeping the questions of error cases unchanged, and require the model to generate standard SQL queries.
We compare the performance with two representative RDB benchmarks, BIRD and Spider 2.0-lite.
As shown in Tab.~\ref{tab:syntax-ablation}, converting TSDBs into RDBs improves EX from 0.00\% to 32.21\%, which is lower than BIRD (56.91\%) but higher than Spider 2.0-lite (22.94\%).
It suggests that the model's poor performance is due not to the difficulty of the questions but to the diverse query syntaxes. 
The model is capable of understanding the questions and generating effective SQL queries, but it lacks specific syntax knowledge, which leads to incorrect answers.
For implementation details, refer to App.~\ref{appendix:ablation-study}.
\section{Conclusion}
In this paper, we introduce \bench, a comprehensive benchmark for evaluating text-to-query capabilities over TSDBs. 
Unlike existing RDB benchmarks, \bench~is designed to capture TSDB-specific challenges by covering diverse query syntaxes, application domains, and time-specific query intents.
Extensive evaluations of advanced LLMs and representative text-to-query methods show that existing methods still struggle with TQTS tasks, particularly in query syntax handling, intent understanding, and schema linking in TSDBs. 
Our analysis reveals key limitations of current methods and highlights opportunities for developing more effective and generalizable text-to-query methods for real-world time-series analysis.

\section*{AI Usage Disclosure}
This work used large language models (LLMs) in three limited capacities: (1) to assess the coverage of manually collected related work; (2) to generate candidate QA pairs during benchmark construction, which were subsequently reviewed and finalized by human experts; and (3) to assist with grammar and phrasing refinement.
LLMs were not used to generate experimental results, figures, or references.
All cited works and final manuscript were verified and approved by the authors, who take full responsibility for the accuracy and integrity of this paper.

\bibliography{iclr2027_conference}
\bibliographystyle{iclr2027_conference}

\appendix
\newpage
\section{Details of \bench}
\label{appendix:tsdb-information}
In this section, we provide detailed information about \bench, including which TSDB management systems were selected for use, what the selected TSDBs are, and which domain they belong to. 
We also describe the query intents covered in \bench, including the identified intent types and their proportions.

\subsection{Identifying TSDB management systems and their query syntax}
\label{appendix:tsdb-ms-and-query-syntax}
We identified 23 TSDB management systems based on their popularity rankings on the TSDB leaderboard and their accessibility, and used them to build \bench. 
Note that several highly ranked TSDB management systems, such as kdb+, were not selected because they require users to apply for and obtain a license before they can be used, which limits their accessibility. 
As shown in Tab.~\ref{tab:tsdb-ms}, we present syntax usage examples for two query intents in each TSDB management system, along with the number of TSDBs included in our benchmark.
Specifically, the examples illustrate how some systems provide
specific functions and operators to help answer questions
involving window aggregation \& resampling (\textbf{I1}) or temporal
change analysis (\textbf{I2}).
The `` / '' indicates that no suitable example was identified for the query interface considered.
\begin{table}[htbp]
\centering
\caption{Selected TSDB management systems and their temporal functions and operators.}
\begin{tabular}{p{0.22\linewidth} p{0.30\linewidth} p{0.26\linewidth}c}
\toprule
\textbf{TSDB management system}
& \multicolumn{2}{c}{\textbf{Representative functions and operators}}
& \textbf{\# TSDB} \\
\cmidrule(lr){2-3}
& \textbf{Window aggregation \newline \& resampling (I1)}
& \textbf{Temporal \newline change analysis (I2)}
& \\
\midrule

\href{https://docs.influxdata.com/influxdb/v2/}{InfluxDB OSS v2}
& \texttt{aggregateWindow()}
& \texttt{derivative()}
& \multirow{2}{*}{17} \\
\href{https://docs.influxdata.com/influxdb3/core/}{InfluxDB 3 Core}
& \texttt{date\_bin\_gapfill()}
& /
& \\

\href{https://prometheus.io/}{Prometheus}
& \texttt{avg\_over\_time()}
& \texttt{resets()}
& 13 \\

\href{https://www.timescale.com/}{TimescaleDB}
& \texttt{time\_bucket\_gapfill()}
& \texttt{delta()}
& 8 \\

\href{https://www.dolphindb.com/}{DolphinDB}
& \texttt{resample()}
& \texttt{ratios()}
& 5 \\

\href{https://druid.apache.org/}{Apache Druid}
& \texttt{TIME\_FLOOR()}
& \texttt{TIMESTAMPDIFF()}
& 4 \\

\href{https://questdb.com/}{QuestDB}
& \texttt{SAMPLE BY}
& \texttt{datediff()}
& 5 \\

\href{https://tdengine.com/}{TDengine}
& \texttt{STATE\_WINDOW()}
& \texttt{CSUM()}
& 4 \\

\href{https://iotdb.apache.org/}{Apache IoTDB}
& \texttt{GROUP BY SESSION()}
& \texttt{TIME\_DIFFERENCE()}
& 4 \\

\href{https://victoriametrics.com/}{VictoriaMetrics}
& \texttt{rollup()}
& \texttt{deriv\_fast()}
& 3 \\

\href{https://basekick.net/}{Arc}
& \texttt{time\_bucket()}
& \texttt{REGR\_SLOPE(v, t)}
& 2 \\

\href{https://griddb.net/en/}{GridDB}
& \texttt{TIME\_SAMPLING()}
& \texttt{TIMESTAMP\_DIFF()}
& 4 \\

\href{https://m3db.io/}{M3DB}
& \texttt{summarize()}
& \texttt{perSecond()}
& 1 \\

\href{https://cratedb.com/}{CrateDB}
& \texttt{date\_bin()}
& \texttt{age()}
& 4 \\

\href{https://www.cnosdb.com/}{CnosDB}
& \texttt{date\_bin()}
& \texttt{idelta\_right()}
& 3 \\

\href{https://arcadedb.com/}{ArcadeDB}
& \texttt{ts.timeBucket()}
& \texttt{ts.delta()}
& 1 \\

\href{https://greptime.com/}{GreptimeDB}
& \texttt{avg\_over\_time()}
& \texttt{increase()}
& 1 \\

\href{https://www.ibm.com/products/db2-event-store}{IBM Db2 Event Store}
& \texttt{window()}
& /
& 4 \\

\href{https://riak.com/products/riak-ts/}{Riak TS}
& /
& /
& 2 \\

\href{https://bangdb.com/}{BangDB}
& \texttt{ROLLUP 5}
& /
& 3 \\

\href{https://www.machbase.com/}{Machbase Neo}
& \texttt{timewindow()}
& \texttt{MAP\_DIFF()}
& 5 \\

\href{https://github.com/4paradigm/OpenMLDB}{OpenMLDB}
& \texttt{ROWS\_RANGE}
& \texttt{drawdown()}
& 3 \\

\href{https://opengemini.github.io/}{openGemini}
& \texttt{GROUP BY time()}
& \texttt{ELAPSED()}
& 1 \\

\bottomrule
\end{tabular}

\label{tab:tsdb-ms}
\end{table}

\subsection{Identifying TSDBs and their domains}
\label{appendix:domain}
\begin{figure}[htbp]
     \centering
     \includegraphics[width=1\linewidth]{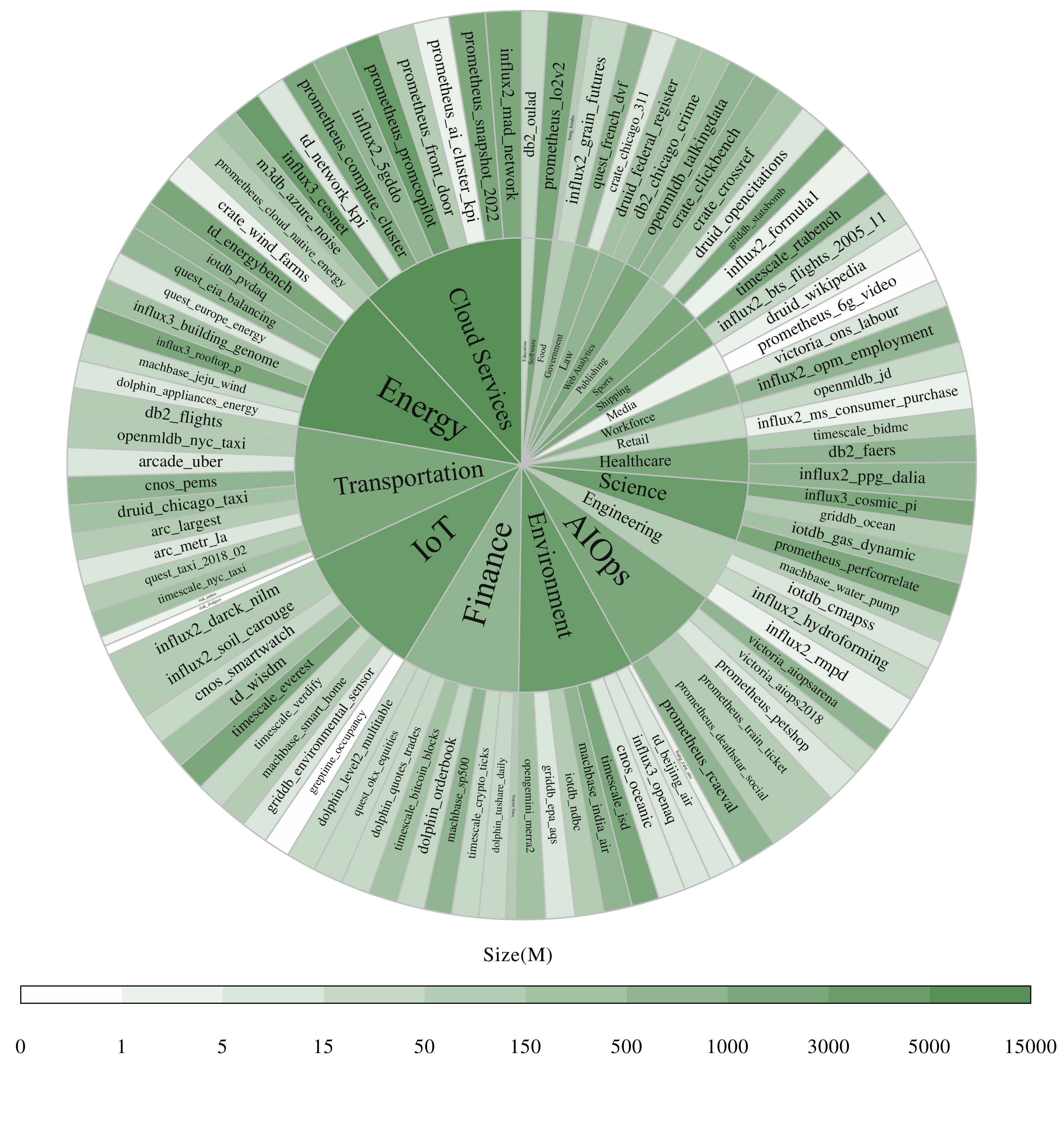}
     \caption{The domain and size of the TSDBs in \bench.}
     \label{fig:tsdb-domain-statistic}
\end{figure}

In \bench, we manually identify a total of 97 high-quality real-world TSDBs for benchmark construction. 
Fig.~\ref{fig:tsdb-domain-statistic} summarizes these TSDBs, along with their application domains, data volumes, and numbers of QA pairs. 
The inner ring denotes domains, while the outer ring denotes individual TSDBs. 
Each sector's angle represents its number of QA pairs, and its color intensity reflects the data volume of the corresponding database: darker shades indicate larger volumes, and lighter shades indicate smaller ones. 
The inner ring covers domains such as cloud services, energy, and IoT, whereas the outer ring presents different TSDBs. 

As shown in Fig.~\ref{fig:tsdb-domain-statistic}, \bench~covers a wide range of real-world domain applications, especially some time-series-intensive domains, such as IoT, AIOps and cloud services.
These domains typically generate and store high-frequency time-series data continuously, resulting in diverse data characteristics and large data volumes.
The total data volume of \bench~is \textbf{41.68 GB}, reflecting the large-scale and diverse nature of real-world time-series workloads collected from various application domains.
To effectively and efficiently analyze these time-series data, many TSDB systems, such as InfluxDB, Prometheus, and TimescaleDB, have been widely adopted. 
The diversity of application domains, together with the different query syntaxes of TSDBs, highlights the significant challenges to real-world TQTS tasks.

\subsection{Details of query intents}
\label{appendix:query-intent}
This subsection details the query intents in \bench, including the identified query intent types, corresponding question examples, and the proportion of each query intent.
\subsubsection{The identified query intents}
We collaborate with domain experts and identify 9 representative query intents in real-world scenarios, including 4 time-specific intents and 5 time-agnostic intents.
The definitions of these query intents, along with their corresponding examples, are presented in Tab.~\ref{tab:query-intents}.
\definecolor{exampleblue}{HTML}{0A74BE}
\definecolor{examplebg}{HTML}{EFF6FC}
\newcommand{\exphl}[1]{%
  \colorbox{examplebg}{\textcolor{exampleblue}{\textbf{#1}}}%
}
\begin{table}[htbp]
\centering
\caption{The definitions and examples of the nine query intents used in \bench.}
\begin{tabular}{
>{\raggedright\arraybackslash}m{0.16\linewidth}
>{\raggedright\arraybackslash}m{0.35\linewidth}
>{\raggedright\arraybackslash}m{0.39\linewidth}}
\toprule

\textbf{Query intent}
& \textbf{Definition}
& \textbf{Example} \\

\midrule
\multicolumn{3}{l}{\textbf{\textit{Time-specific}}} \\
\midrule

\textbf{I1}: Window aggregation \& resampling
&
Resample or aggregate a time series over time windows or a new sampling grid, yielding results for each window or grid.
&
``What was Hankyung's average temperature forecast \exphl{for each day} of June 2017?'' \\


\textbf{I2}: Temporal change analysis
&
Analyzing temporal changes in values within a single time series.
&
``What was the largest \exphl{one-second increase} in whole-home active power?'' \\


\textbf{I3}: Time localization
&
Locate the time point(s) or interval at which an event or state occurs.
&
``\exphl{When} was the earliest weather observation in the database?'' \\


\textbf{I4}: Relationship analysis
&
Analyzing relationships between two or more independently identifiable time series.
&
``On September 12, at which minutes did \exphl{Apple's trading volume exceed} \exphl{Amazon's}?'' \\

\midrule
\multicolumn{3}{l}{\textbf{\textit{Time-agnostic}}} \\
\midrule

\textbf{I5}: Conditional filtering
&
Filtering samples or events based on specified conditions.
&
``Return the full records for matches played \exphl{after June 15, 2024}.'' \\


\textbf{I6}: Statistical computation
&
Computing statistical values from samples or events.
&
``What was the \exphl{average fare across all trips}?'' \\


\textbf{I7}: Sort \& rank
&
Ranking candidates by one or more ordering criteria.
&
``Show the \exphl{first 100} sensor readings \exphl{in chronological order}.'' \\


\textbf{I8}: Metadata query
&
Querying metadata about time-series databases, such as schema information.
&
``How is \exphl{each column} in the flight-record table \exphl{defined} in Db2?'' \\


\textbf{I9}: Domain knowledge dependency
&
Requiring the use of specialized domain knowledge to answer the question.
&
``What were the 15-minute \exphl{OHLC} \exphl{prices} and traded share volume for each equity on March 27, 2025?'' \\

\bottomrule
\end{tabular}

\label{tab:query-intents}
\end{table}


\subsubsection{Statistic of query intents}
\begin{figure}[htbp]
     \centering
     \includegraphics[width=1\linewidth]{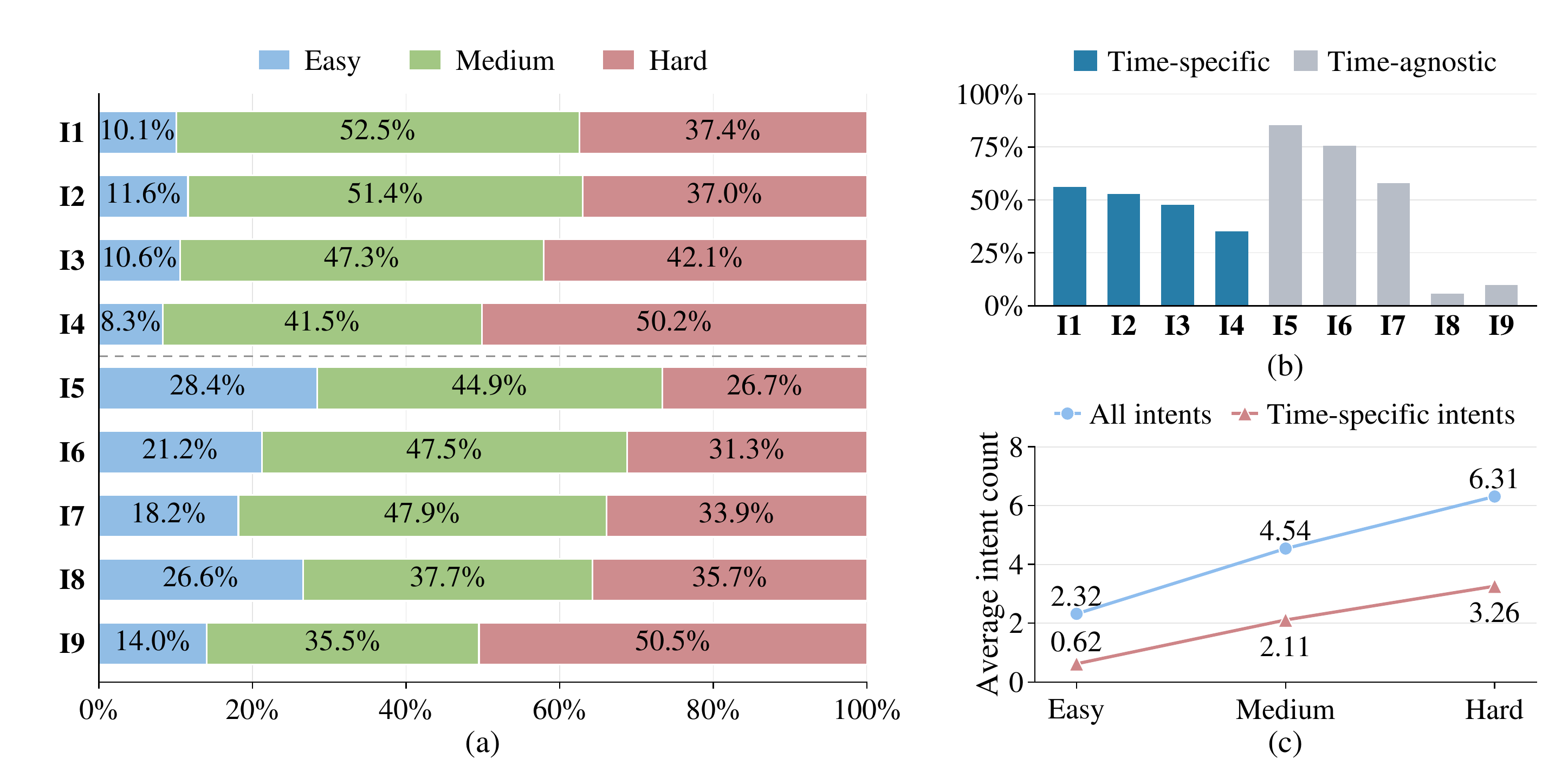}
     \caption{The statistics of query intents. (a) Distribution of query intents across different difficulty levels; (b) Distribution of query intents in \bench; (c) Average query intent count across different difficulty levels.}
     \label{fig:intent_statistics}
\end{figure}

Fig.~\ref{fig:intent_statistics} presents the statistics of query intents. 
To examine whether the distribution of query intents is unbiased with respect to query difficulty, we analyze the proportion of each intent across different difficulty levels, as shown in Fig.~\ref{fig:intent_statistics}(a). 
Overall, each query intent is relatively evenly distributed across different difficulty levels, suggesting that query difficulty is not dominated by any particular intent type and that the benchmark maintains a balanced intent distribution across difficulty levels.
Nevertheless, time-specific intents (\textbf{I1}--\textbf{I4}) occur more frequently in hard questions, and less in easy questions, indicating that temporal reasoning tends to introduce additional complexity.

Fig.~\ref{fig:intent_statistics}(b) further shows that the overall distribution of query intents is consistent with real-world information-seeking demands. 
For example, \textbf{I5: }\emph{Conditional Filtering} and \textbf{I6: }\emph{Statistical Computation} account for the largest proportions, reflecting their prevalence in practical querying scenarios.
We additionally investigate how the number of query intents varies with query difficulty, as shown in Fig.~\ref{fig:intent_statistics}(c). 
As the difficulty increases, queries tend to involve more intents, with time-specific query intents showing the most pronounced increase.

\subsection{Non-unified schema structures across TSDBs}
\label{appendix:diverse-schema}
\begin{figure}[t]
     \centering
     \includegraphics[width=1\linewidth]{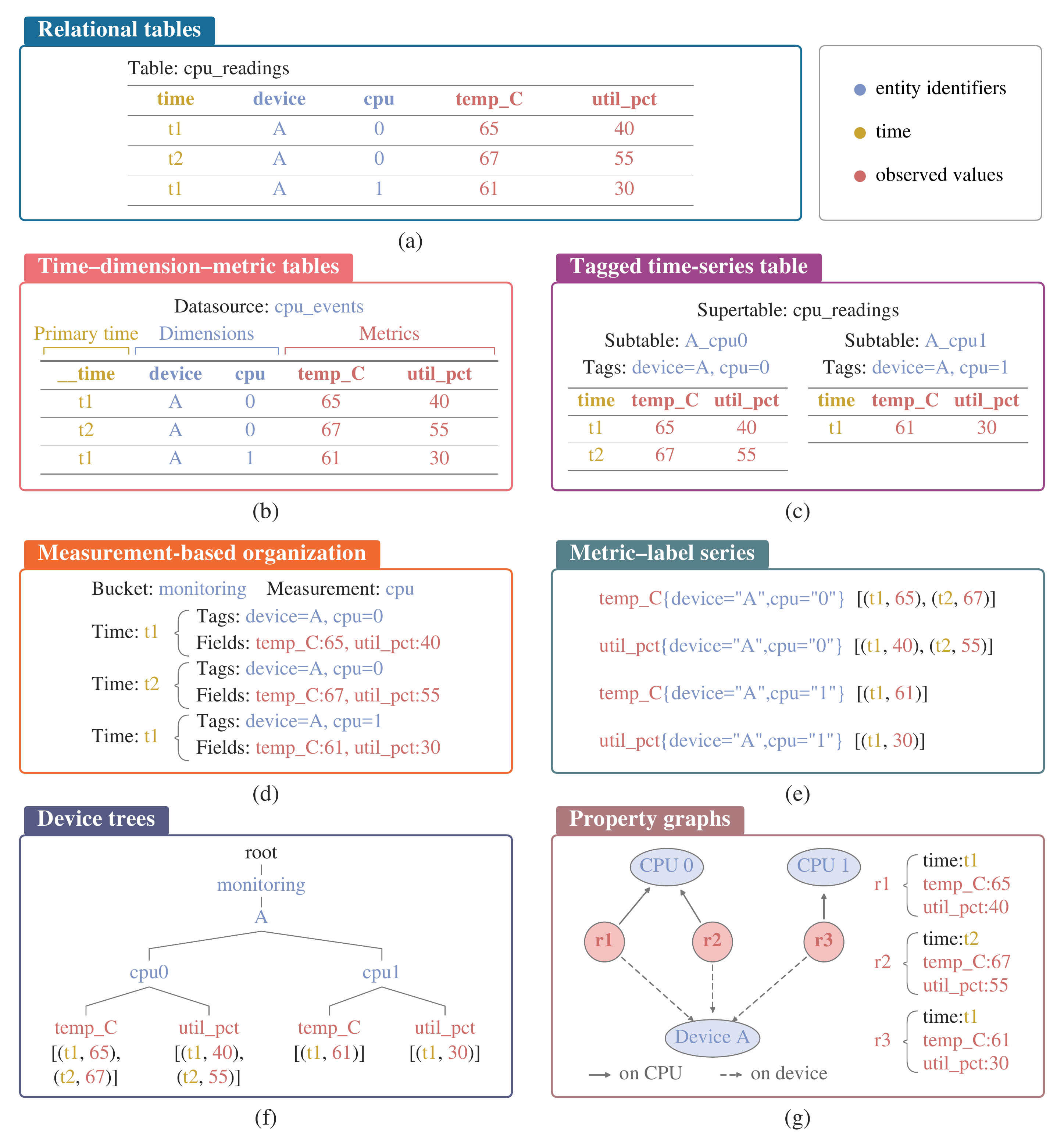}
     \caption{The seven types of schema structures of the same time-series data.}
     \label{fig:diverse-schema-framed}
\end{figure}

Here, we illustrate the non-unified schema structures of TSDBs.
TSDB schemas describe the logical organization of time-series data and can be broadly grouped into seven categories:
\textbf{Relational tables}, \textbf{Time–dimension–metric tables}, \textbf{Tagged time-series table}, \textbf{Measurement-based organization}, \textbf{Metric–label series}, \textbf{Device tree} and \textbf{Property graphs}.
Fig.~\ref{fig:diverse-schema-framed} shows how the same time-series data are organized differently across these seven schema structures.
The descriptions of these non-unified seven schema structures are as follows:
\begin{itemize}
    \item \textbf{Relational tables} store records in named tables with typed columns. Each row contains attributes such as entity identifiers, timestamps, and values. Each table has a primary key that uniquely identifies records, while some columns may serve as foreign keys to reference records in related tables.
    
    \item \textbf{Time--dimension--metric tables} store event records, where each row represents an event associated with a datasource. Each row contains a primary timestamp, dimensions describing event attributes, and metrics representing quantitative values for analysis.
    
    \item \textbf{Tagged time-series tables} store timestamped observations, where each row represents an observation associated with an entity or series. Each row contains tags for identifying the entity or series and data columns recording observed values.
    
    \item \textbf{Measurement-based organization} stores observation points under different measurements within a data container. Each point contains a timestamp, tags describing the observation context, and fields recording measured values.
    
    \item \textbf{Metric--label series} represent time-series data as collections of timestamped samples. Each series contains a metric name, a set of labels identifying the series, and samples recording the metric values over time.
    
    \item \textbf{Device trees} store device and measurement data in a hierarchical structure. Each root-to-leaf path represents a time series, and each point along the path contains a timestamp and an observed value.
    
    \item \textbf{Property graphs} represent entities and observations as vertices connected by edges. Each vertex and edge contain properties that describe attributes such as timestamps and observed values.
\end{itemize}

\subsection{Details of the TSDB context}
For each TSDB included in \bench, we provide a customized context that captures its background information, key characteristics, and concrete usage scenarios. 
By incorporating this contextual knowledge, the model can gain a deeper understanding of the target TSDB and its application requirements, allowing it to generate questions that better reflect real-world TSDB usage scenarios and business needs.  

For example, as shown in Fig.~\ref{fig:prometheus_context}, the context of a Prometheus database contains descriptions of its background, including the fixed benchmark time setting, and the real-world cloud-service telemetry scenarios involving Kubernetes, node, application, Loki, PostgreSQL, and service metrics.
\begin{figure}[t]
    \centering
    \begin{promcopilotbox}{Database context used in \bench~construction process}
    \begin{lstlisting}[basicstyle=\ttfamily\small]
{
  "name": "Prometheus March 2022 cloud-service snapshot",
  "description": "Seven contiguous native Prometheus blocks 
  containing real Kubernetes, node, application, Loki, PostgreSQL, 
  and service telemetry from a cloud-service environment.",
  "engine": {
    "name": "prometheus",
    "version": "2.54.1"
  },
  "query_language": "PromQL: Prometheus Query Language for instant
  queries at the fixed benchmark current_time, using declared metric 
  and label names. Guidance: The query language is PromQL, and 
  Prometheus has no SQL tables.",
  "time": {
    "current_time": "2022-03-23T17:59:59Z",
    "timezone": "UTC",
    "ranges": [
      {
        "start": "2022-03-18T12:00:00Z",
        "end": "2022-03-23T18:00:00Z"
      }
    ]
  }
    \end{lstlisting}
    \end{promcopilotbox}
    \caption{An example of the customized Prometheus database context used in \bench.}
    \label{fig:prometheus_context}
\end{figure}

\section{Details of QA construction}
\label{appendix:interface}
\subsection{Design of visual analytics annotation tool}



Candidate QA pairs constructed with AI assistance may contain incorrect queries, unnatural wording, or mismatches between the intended request and the database context.
To support systematic human verification before benchmark inclusion, we developed a visual analytics tool guided by three quality requirements.
These requirements concern answer correctness, question naturalness and clarity, and alignment with query intents and TSDB context.

\noindent\textbf{R1: Ensuring the correctness of executable answers.}
Reliable answers are essential for evaluating model performance, yet successful execution does not establish whether a query correctly answers its associated question.
We therefore require independent human interpretations and execution evidence to assess answer correctness beyond execution success.

\noindent\textbf{R2: Ensuring the naturalness and clarity of questions.}
AI-assisted construction requires explicit review of whether generated questions express meaningful requests in natural, comprehensible language.
Awkward phrasing, ambiguity, and mechanical descriptions can obscure the intended meaning.
The system must therefore help reviewers identify these issues and support targeted refinement before benchmark inclusion.

\noindent\textbf{R3: Ensuring alignment with query intents and TSDB context.}
A fluent question and an executable answer do not necessarily constitute a suitable benchmark example.
Each pair must faithfully represent the selected query intents, reflect a meaningful TSDB scenario, and be answerable from the supplied context.
Human review must therefore identify misalignment, missing information, and unstated assumptions.

\begin{figure}[htbp]
    \centering
    \includegraphics[width=1\linewidth]{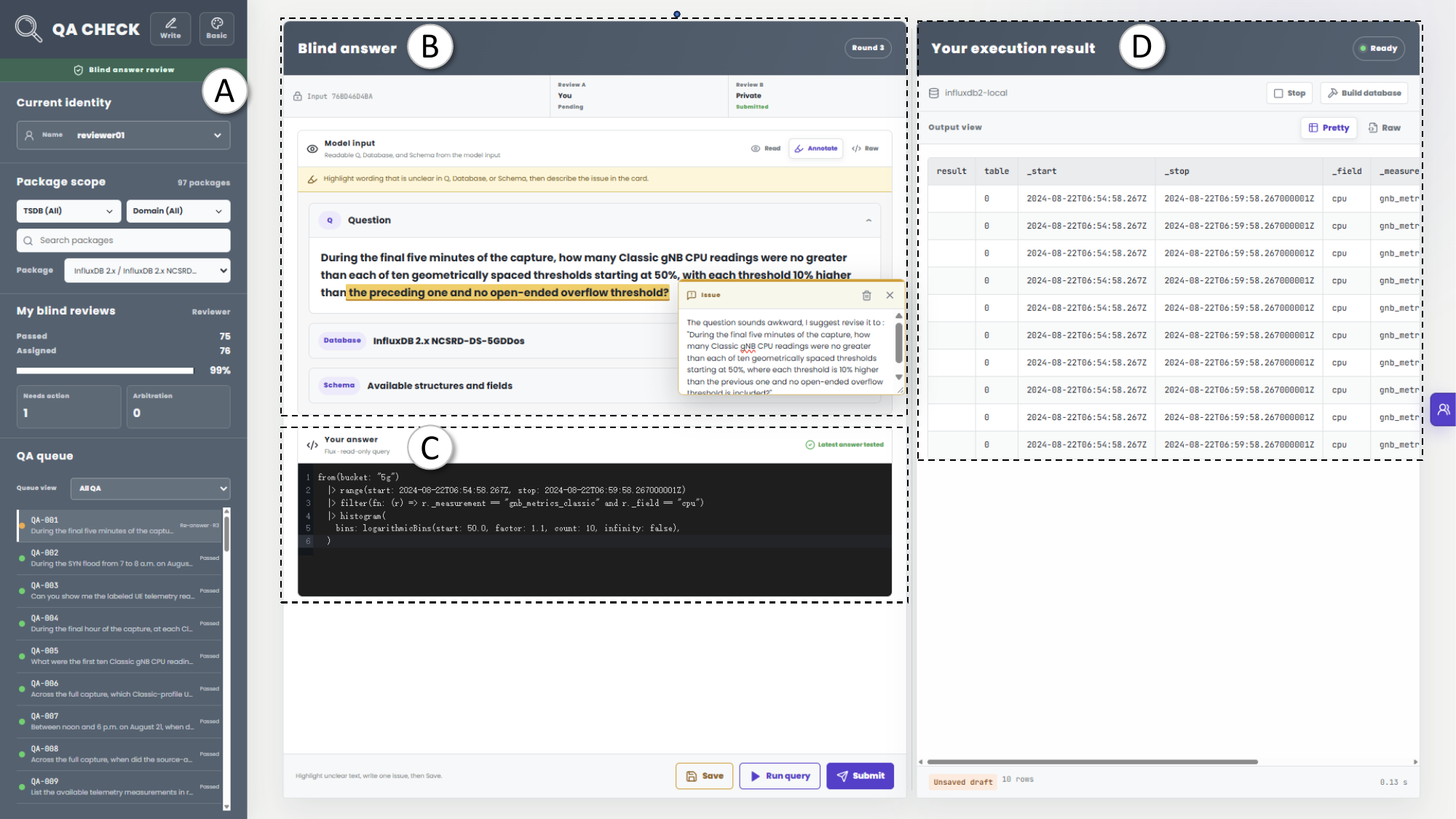}
    \caption{The visual interface for annotators in human-AI collaboration step.}
    \label{fig:reviewer}
\end{figure}

\begin{figure}[htbp]
    \centering
    \includegraphics[width=1\linewidth]{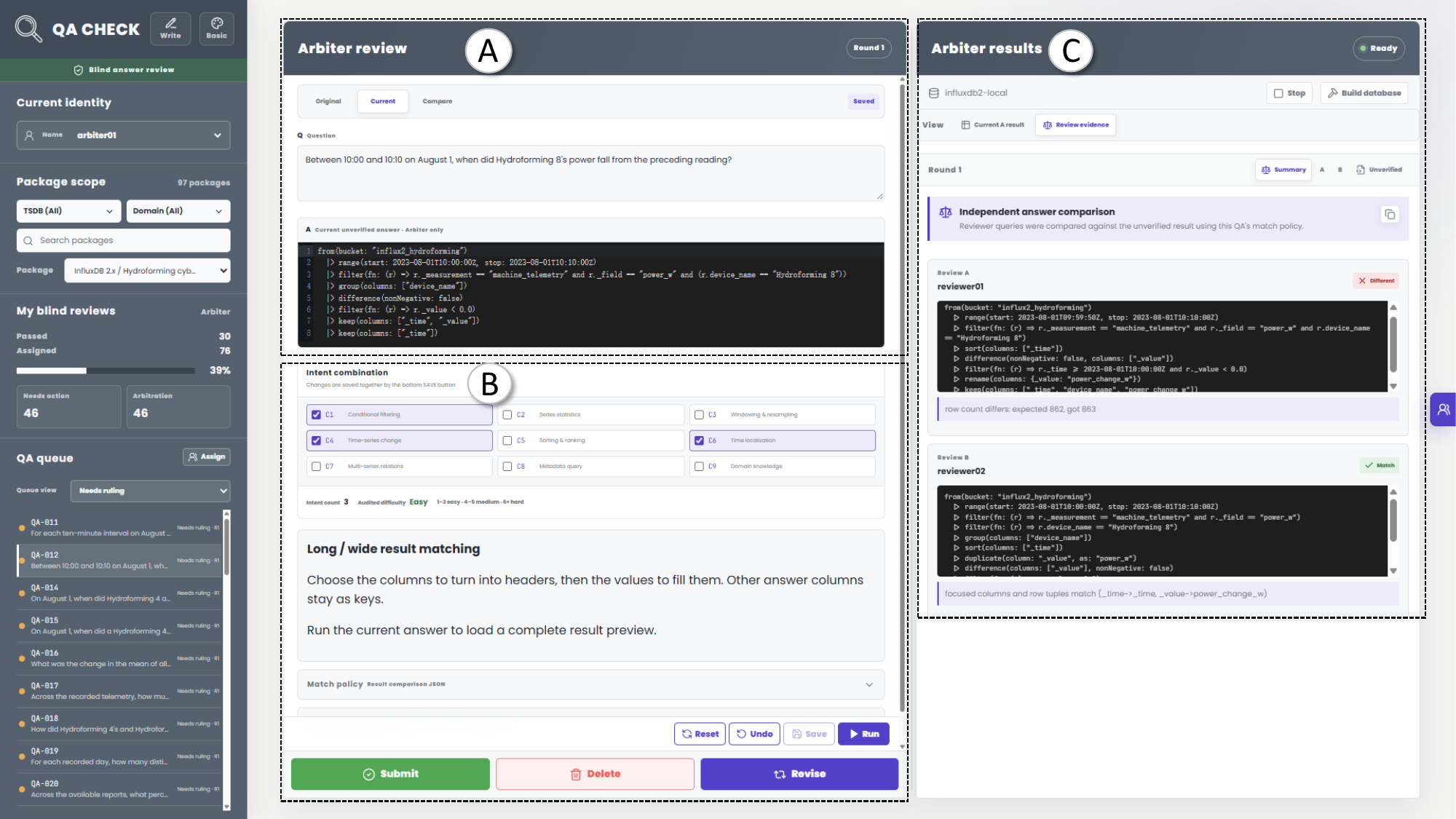}
    \caption{The visual interface for adjudicator in QA verification step.}
    \label{fig:arbiter}
\end{figure}


\subsection{Detail of QA verification}
The verification workflow begins with two annotators independently reviewing each candidate QA pair.
Fig.~\ref{fig:reviewer}A enables each annotator to select TSDB and inspect current progress.
Fig.~\ref{fig:reviewer}B presents the question, database description, and schema while hiding the candidate answer and the peer annotator's solution.
Annotators manually construct their own queries in the editor (Fig.~\ref{fig:reviewer}C), providing independent answers to the questions and examining whether the available context supports answering it.
The annotation interface in Fig.~\ref{fig:reviewer}B allows them to flag unclear wording or insufficient information.
They can then inspect their queries' execution results and refine their answers before submission (Fig.~\ref{fig:reviewer}D).

When verification reveals inconsistent results or an unanswerable question, an adjudicator examines the QA pair together with its intent combination, match policy, and annotator feedback.
Fig.~\ref{fig:arbiter} shows the interface for adjudicator to review and revise the unverified QA pair.
The adjudicator can inspect the QA pair (Fig.~\ref{fig:arbiter}A), its corresponding query intents and can check the query result (Fig.~\ref{fig:arbiter}B).
The Review evidence view brings together both annotator queries and execution results for comparison (Fig.~\ref{fig:arbiter}C).
These views support assessment of query correctness alongside the question's wording, intended meaning, and contextual requirements.

The adjudicator revises the QA pair based on the identified issues, after which it undergoes another verification round.
This process continues until both annotators independently obtain results consistent with the revised answer.
The tool thus supports human review and correction of AI-assisted construction outputs, keeping interpretation, revision, and verification under human control.

The tool also records active review time and review-related actions, including query executions, annotations, review decisions, and arbitration operations.
The Human activity dashboard summarizes these records across database packages and displays review progress.
These records allow us to document the human effort involved in QA verification and provide the basis for the verification effort reported in the main text.

\section{More experiment detail and analysis}
\label{appendix:more-analysis}

\subsection{Evaluation criteria for \bench}
\label{appendix:evaluation-criteria}

Here, we introduce the detailed criteria for the Execution Accuracy (\textbf{EX}) metric used in the experiment for \bench.
Inspired by Spider 2.0 \citep{lei2025spider2}, EX evaluates whether a predicted query produces the expected results, primarily by checking whether its output contains all columns returned by the gold query.
However, in TSDBs, this situation becomes more complex, since models can often generate different queries that return the same underlying data in different formats, such as queries involving the \texttt{pivot} function. 

\begin{figure}[htbp]
    \centering
    \includegraphics[width=1\linewidth]{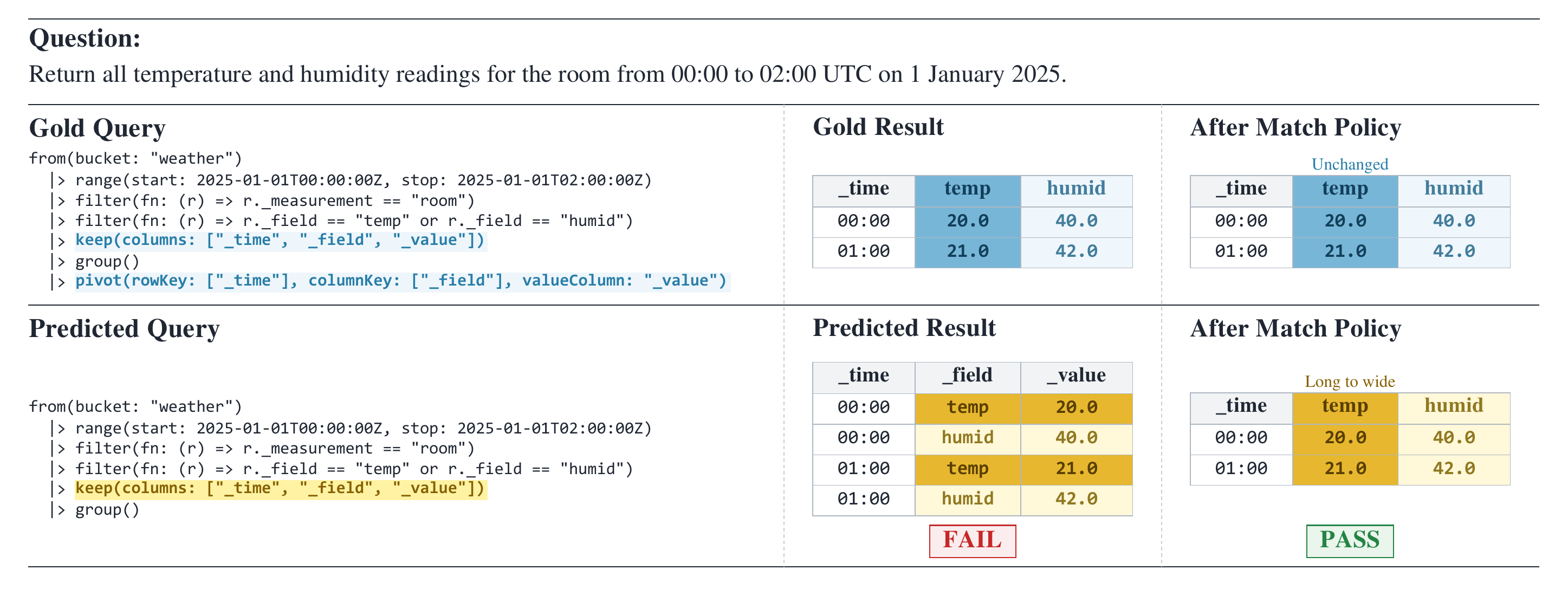}
    \caption{The evaluation criteria and an example of the match policy used in the evaluation.}
    \label{fig:ex-match-policy}
\end{figure}

As illustrated in Fig.~\ref{fig:ex-match-policy}, the absence of the \texttt{pivot} function results in a different output format between the predicted and gold results, causing the evaluation to incorrectly mark the prediction as incorrect. 
To address this issue, inspired by the matching policy in Spider 2.0, we introduce an additional transformation step to establish a more precise and suitable evaluation criterion for TQTS tasks.
This policy reduces false negatives in evaluation while avoiding an increase in false positives.

\subsection{Detailed Information for Human Evaluation}
\label{appendix:human-evaluation}
In the human evaluation, we utilize the developed visual analytics tool as the evaluation interface. 
Specifically, human experts are provided with the question, database schema, and TSDB context, and are asked to manually write the corresponding queries. 
After completion, we evaluate the EX of the queries provided by human experts.

 \subsection{WHY THE RDB METHODS STRUGGLE WITH THE TQTS TASK}
\label{appendix:rdb-methods}
The experimental results demonstrate that existing RDB methods generally perform poorly on the TQTS task. 
To investigate the reasons behind their limitations, we select DeepEye-SQL, a representative and high-performing text-to-query method on BIRD, for further analysis.

For text-to-query tasks, DeepEye-SQL consists of four main stages: schema linking, query generation, query verification and refinement, and confidence-based selection. 
However, these stages are primarily designed for generating SQL queries.
Specifically, as shown in Fig.~\ref{fig:deepeye-syntax-dependence}, the prompts used in the query generation stage are heavily constrained by SQL syntax. 
Such SQL-specific guidance limits the model’s ability to generate queries required by TSDBs, which often adopt different query syntaxes and syntactic structures.

As shown in Fig.~\ref{fig:deepeye-promql-sql-case}, the generated queries may not conform to the syntax requirements of TSDB query syntax, leading to execution failures. 
Therefore, the SQL-specific design of existing RDB methods is a key reason why they struggle with the TQTS task, as it prevents them from effectively adapting to the diverse query syntaxes of TSDBs.

\begin{figure}[htbp]
    \centering
    \begin{deepeyebox}{DeepEye-SQL: the original skeleton-based generation prompt}
    \textbf{\# Task:}\par
    You are an expert \textbf{SQL developer} who uses a systematic approach to generate complex \textbf{SQL queries}.
    Your task is to analyze the given question and database schema, then generate a \textbf{SQL query} using a three-step process:\par
    {[\ldots]}

    \textbf{\#\# Step 1: Plan (SQL Components Analysis)}\par
    Analyze the question and identify:\par
    - \textbf{SELECT clause}: What data needs to be retrieved? (columns, aggregations, calculations)\par
    - \textbf{FROM clause}: Which tables are needed?\par
    - \textbf{JOIN clauses}: What relationships need to be established?\par
    - \textbf{WHERE clause}: What filtering conditions are required?\par
    - \textbf{GROUP BY clause}: What grouping is needed for aggregations?\par
    {[\ldots]}\par
    - \textbf{ORDER BY clause}: What sorting is required?\par
    {[\ldots]}

    \textbf{\# Important Rules:}\par
    {[\ldots]}\par
    2. \textbf{SQLite Compatibility: Use only SQLite-compatible functions and syntax}\par
    {[\ldots]}\par
    \textbf{\# Output Format:}\par
    Please respond with XML code structured as follows:\par
    {[\ldots]}\par
    \texttt{<result>}\par
    The final \textbf{SQL query} that answers the target question and \textbf{can be executed on the target SQLite database}, ensure there is not any SQLite comment and not any other explanation text in the SQL query.\par
    {[\ldots]}\par
    \texttt{</result>}\par
    {[\ldots]}
    \end{deepeyebox}

    \caption{An excerpt from the SQL generation prompt of DeepEye-SQL.}
    \label{fig:deepeye-syntax-dependence}
\end{figure}

\begin{figure}[htbp]
    \centering
    \begin{deepeyefailurebox}{Failure case: PromQL required, SQL generated}
    \textbf{Question}\par
    Which streamer containers are currently in the running state during the 200 MiB three-streamer test?

    \textbf{Gold query (PromQL)}
\begin{lstlisting}[style=deepeyequery]
kube_pod_container_status_running{
  dataset="prom_6g_video",
  bandwidth_mib="200",
  streamer_count="3"
} > 0
\end{lstlisting}
    \textbf{DeepEye predicted query (SQL)}
\begin{lstlisting}[style=deepeyequery]
SELECT DISTINCT k.streamer
FROM kube_pod_container_status_running AS k
WHERE k.bandwidth_mib = '200'
  AND k.streamer_count = '3' AND k._value = 1
\end{lstlisting}
    \end{deepeyefailurebox}
    \caption{An example where DeepEye-SQL generates SQL instead of the required PromQL.}
    \label{fig:deepeye-promql-sql-case}
\end{figure}

\subsection{Why the TSDB methods still struggle with TQTS task}
\label{appendix:tsdb-methods}
In the experiment, TSDB methods still achieve limited EX on \bench.
To understand the reasons behind such poor performance, we conduct a further analysis of the state-of-the-art method, PromCopilot, a text-to-query method designed for Prometheus-based monitoring scenarios.

PromCopilot addresses the TQTS task through two stages.
First, it constructs a knowledge graph to describe system contexts and facilitate schema linking.
However, this knowledge graph is tightly coupled with Kubernetes cloud-native environments, where entities and relations are defined based on components such as pods, nodes, and services.
Such a domain-specific schema makes it difficult to directly transfer the method to other domains.
Second, PromCopilot generates queries using LLMs with carefully designed prompts. 

\begin{figure}[htbp]
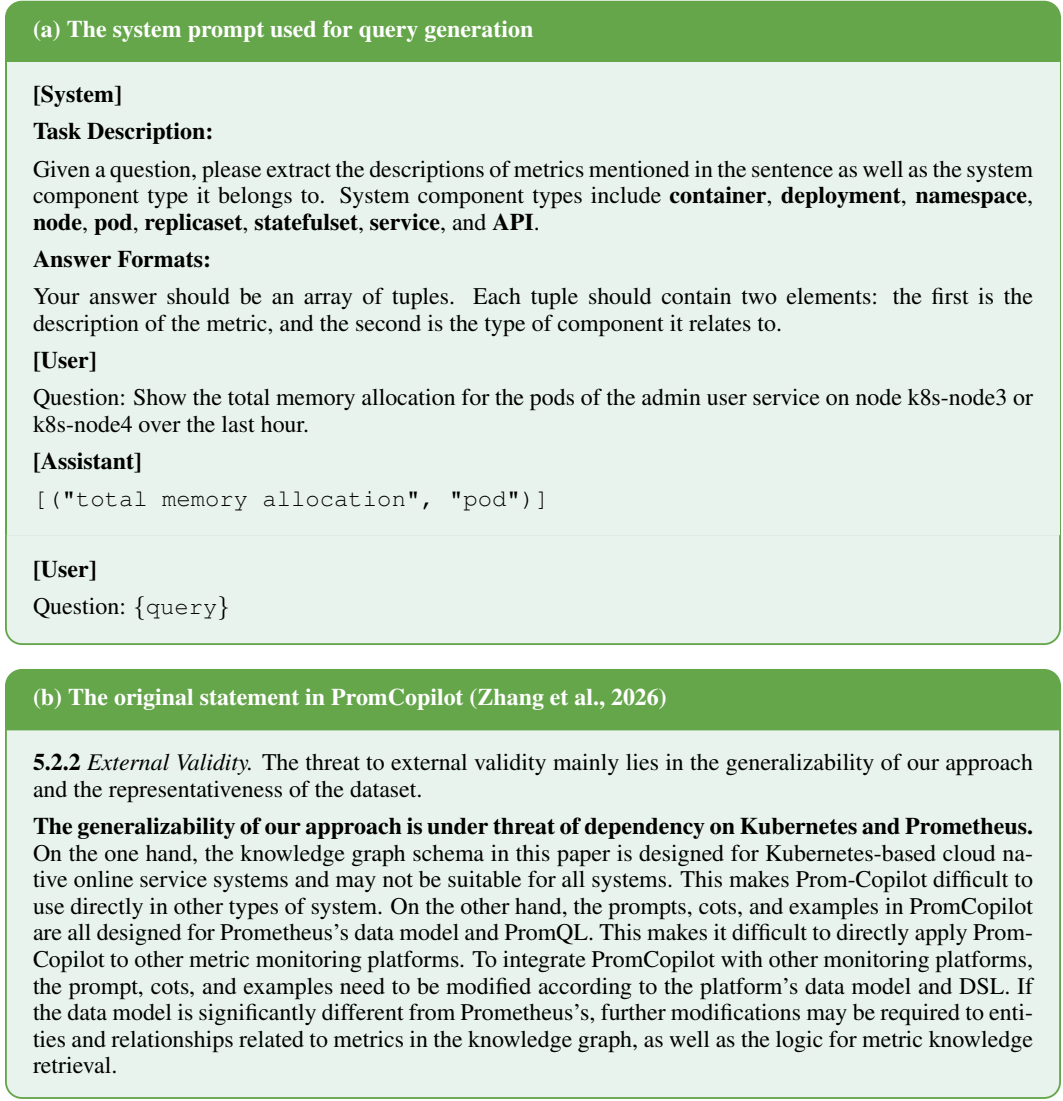

    \centering
    \begin{promcopilotbox}{(a) The system prompt used for query generation}
    \textbf{[System]}\par
    \textbf{Task Description:}\par
    Given a question, please extract the descriptions of metrics mentioned in the sentence as well as the system component type it belongs to. System component types include \textbf{container}, \textbf{deployment}, \textbf{namespace}, \textbf{node}, \textbf{pod}, \textbf{replicaset}, \textbf{statefulset}, \textbf{service}, and \textbf{API}.
    
    \textbf{Answer Formats:}\par
    Your answer should be an array of tuples. Each tuple should contain two elements: the first is the description of the metric, and the second is the type of component it relates to.
    
    \textbf{[User]}\par
    Question: Show the total memory allocation for the pods of the admin user service on node k8s-node3 or k8s-node4 over the last hour.
    
    \textbf{[Assistant]}\par
    \texttt{[("total memory allocation", "pod")]}
    
    \textbf{[User]}\par
    Question: \texttt{\{query\}}
    \end{promcopilotbox} 
    \begin{promcopilotbox}{(b) The original statement in PromCopilot~\citep{zhang2026promcopilot}}
    \textbf{5.2.2} \textit{External Validity.}
    The threat to external validity mainly lies in the generalizability of our approach and the representativeness of the dataset.
    
    \textbf{The generalizability of our approach is under threat of dependency on Kubernetes and Prometheus.}
    On the one hand, the knowledge graph schema in this paper is designed for Kubernetes-based cloud native online service systems and may not be suitable for all systems. This makes Prom-Copilot difficult to use directly in other types of system. On the other hand, the prompts, cots, and examples in PromCopilot are all designed for Prometheus's data model and PromQL. This makes it difficult to directly apply PromCopilot to other metric monitoring platforms. To integrate PromCopilot with other monitoring platforms, the prompt, cots, and examples need to be modified according to the platform's data model and DSL. If the data model is significantly different from Prometheus's, further modifications may be required to entities and relationships related to metrics in the knowledge graph, as well as the logic for metric knowledge retrieval.
    
    \end{promcopilotbox}

    \caption{The description of tightly coupled domains as reflected in 
    (a) the prompt for query generation and 
    (b) the statement of external validity in the original paper.}
    \label{fig:promcopilot-domain-dependence}
\end{figure}

As illustrated in Fig.~\ref{fig:promcopilot-domain-dependence}(a), the system prompts are primarily tailored for cloud services and AIOps scenarios, further limiting their applicability beyond the original domain.
These domain-specific designs hinder the generalization of PromCopilot to broader TSDB application domains, which may explain its limited performance on \bench, where it achieves only 0.69\% EX. 
This observation is also consistent with the authors' acknowledgment that PromCopilot relies on Kubernetes- and Prometheus-specific knowledge (Fig.~\ref{fig:promcopilot-domain-dependence}(b)).

\newpage
\subsection{More Examples of Error Analysis}
\label{appendix:error-categories-examples}
We provide additional examples for each error type identified in error analysis. 
Tab.~\ref{tab:error-category-examples} summarizes the error taxonomy and links each 
subcategory to representative error cases. 
The following figures present detailed examples, 
illustrating how these errors manifest in generated queries and how they affect the 
correctness of the final results.
Here, for intent understanding errors, we only consider the errors with time-specific query intents.

\begin{table}[htbp]
\renewcommand{\arraystretch}{1.5}
\centering
\caption{Overview of the error types and their representative examples.}
\label{tab:error-category-examples}

\begin{tabular}{lll}
\hline
\textbf{Error Type}   
& \textbf{Subcategory} 
& \textbf{Representative Example(s)}    
\\ \toprule
\multirow{2}{*}{Query syntax errors}         
& Query structure errors    
& Refer to Fig.~\ref{fig:error_syntax_case_study}(a) and Fig.~\ref{fig:appendix-errors-query-syntax}(a) \\ \cline{2-3} 
 & Function/keyword usage errors & Refer to Fig.~\ref{fig:error_syntax_case_study}(b) and Fig.~\ref{fig:appendix-errors-query-syntax}(b)\\ \midrule
\multirow{4}{*}{Intent understanding errors} & \begin{tabular}[c]{@{}l@{}}\textbf{I1}: Window aggregation \\      \& resampling\end{tabular} & Refer to Fig.~\ref{fig:intent_errors_combined}(b)   \\ \cline{2-3} 
& \textbf{I2}: Temporal change analysis  & Refer to Fig.~\ref{fig:appendix-errors-intent-understanding}(a)                 \\ \cline{2-3} 
& \textbf{I3}: Time localization      & Refer to Fig.~\ref{fig:appendix-errors-intent-understanding}(b)                 \\ \cline{2-3} 
& \textbf{I4}: Relationship analysis  & Refer to Fig.~\ref{fig:appendix-errors-intent-understanding}(c)                 \\ \midrule
Schema linking errors  &  / & Refer to Fig.~\ref{fig:appendix-errors-schema-linking}(a) and Fig.~\ref{fig:appendix-errors-schema-linking}(b)  \\ \bottomrule
\end{tabular}
\end{table}

\begin{figure}[htbp]
    \centering
    \includegraphics[width=1\linewidth]{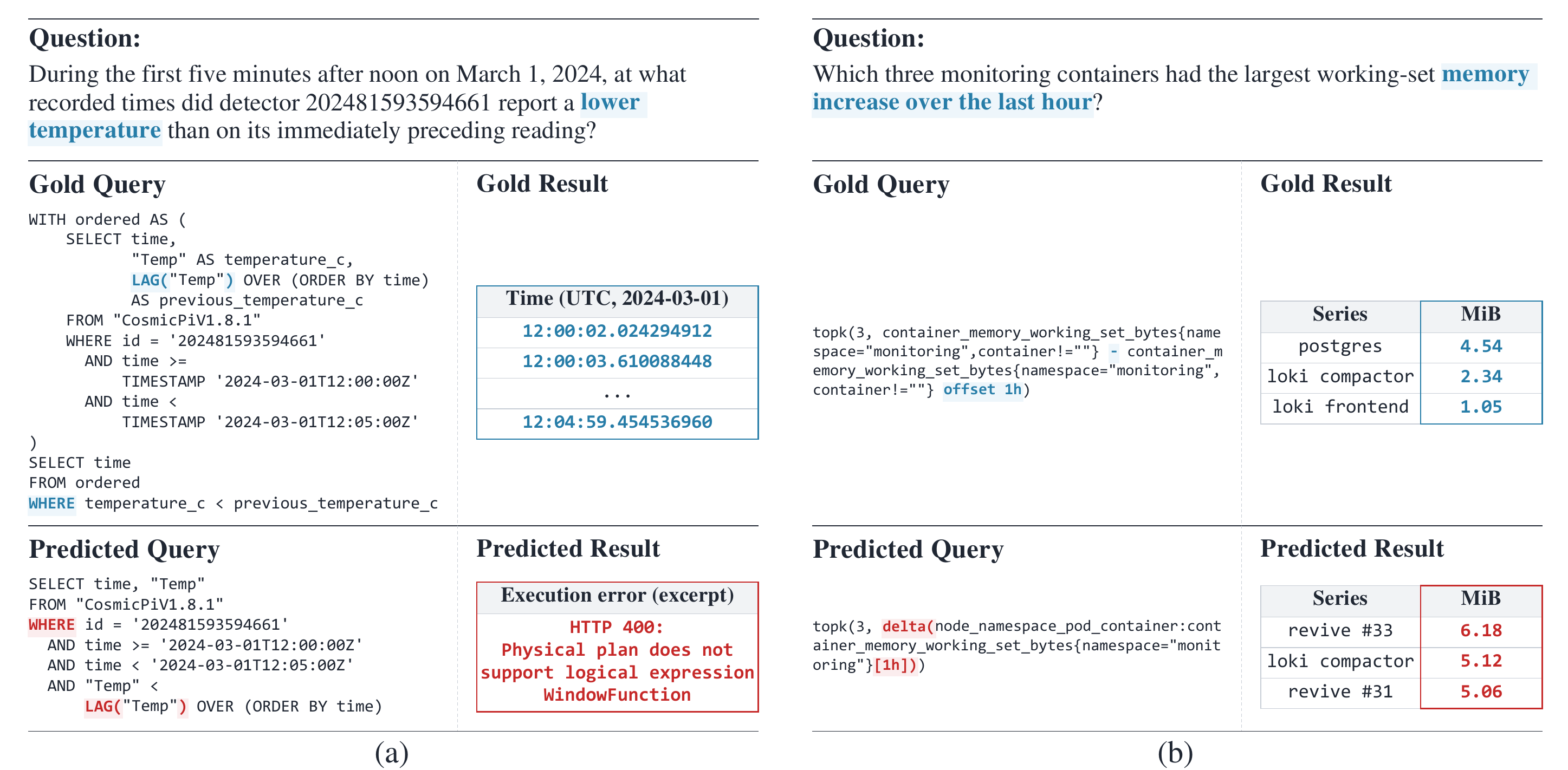}
    \caption{Examples of query syntax errors. (a) Query structure error: placing \texttt{LAG()} directly in the \texttt{WHERE} clause, rather than computing the preceding reading in a subquery, causes a window-function execution error. (b) Operator usage error: replacing subtraction with \texttt{offset 1h} by \texttt{delta(...[1h])} substitutes an extrapolated change for the difference between two time points; together with a changed metric, this produces incorrect values and rankings.}
    \label{fig:appendix-errors-query-syntax}
\end{figure}

\begin{figure}[htbp]
    \centering
    \includegraphics[width=1\linewidth]{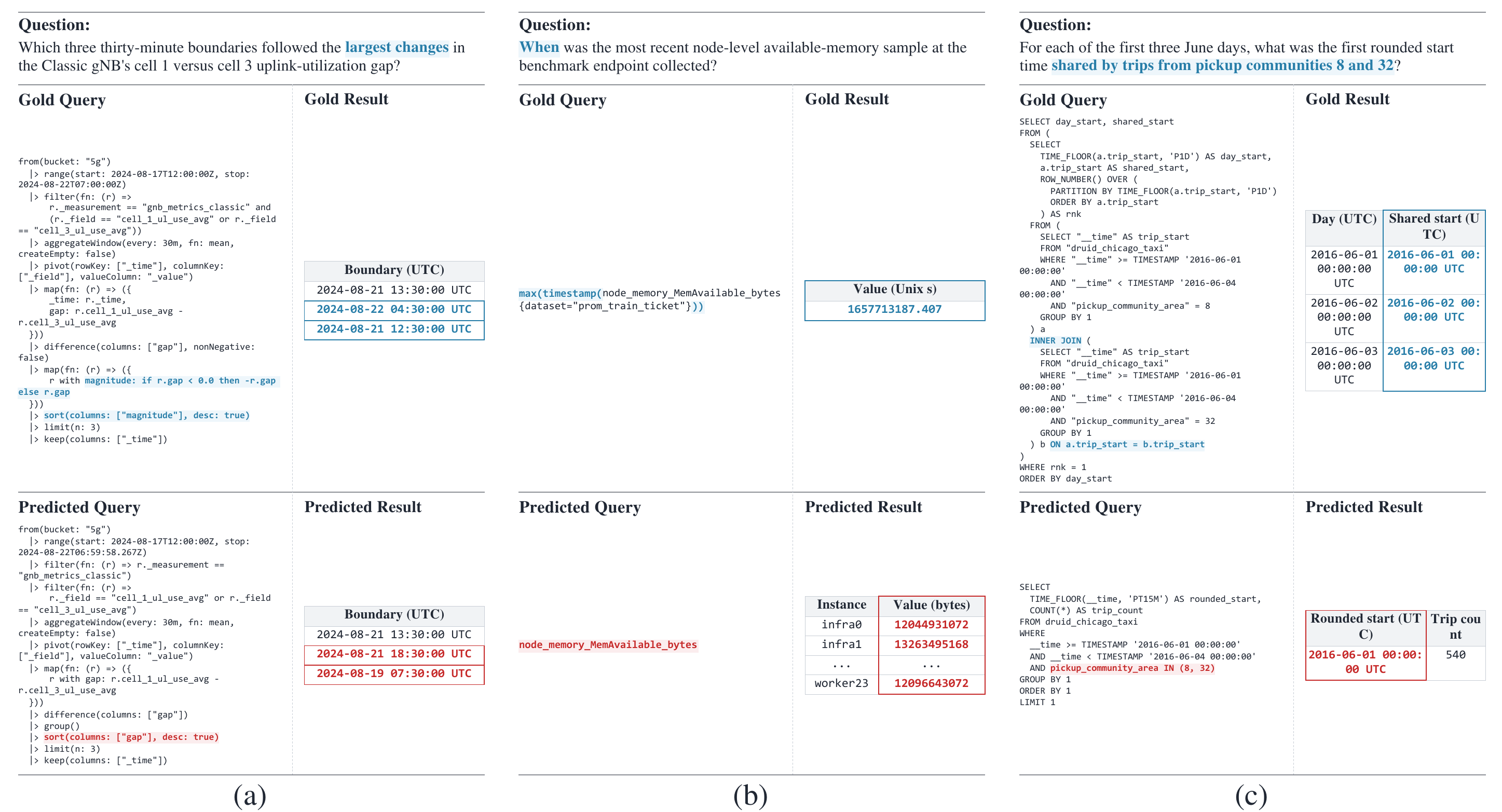}
    \caption{Examples of intent understanding errors. (a) Temporal change analysis error: ranking signed differences rather than absolute magnitudes interprets the largest changes as the largest increases, selecting incorrect time boundaries. (b) Time localization error: returning available-memory values instead of applying \texttt{max(timestamp(...))} reports memory quantities rather than the latest sample time. (c) Relationship analysis error: using \texttt{IN (8, 32)} instead of matching timestamps across both communities confuses union with intersection; a global \texttt{LIMIT 1} further returns a single time rather than the earliest shared time for each day.}
    \label{fig:appendix-errors-intent-understanding}
\end{figure}

\begin{figure}[htbp]
    \centering
    \includegraphics[width=1\linewidth]{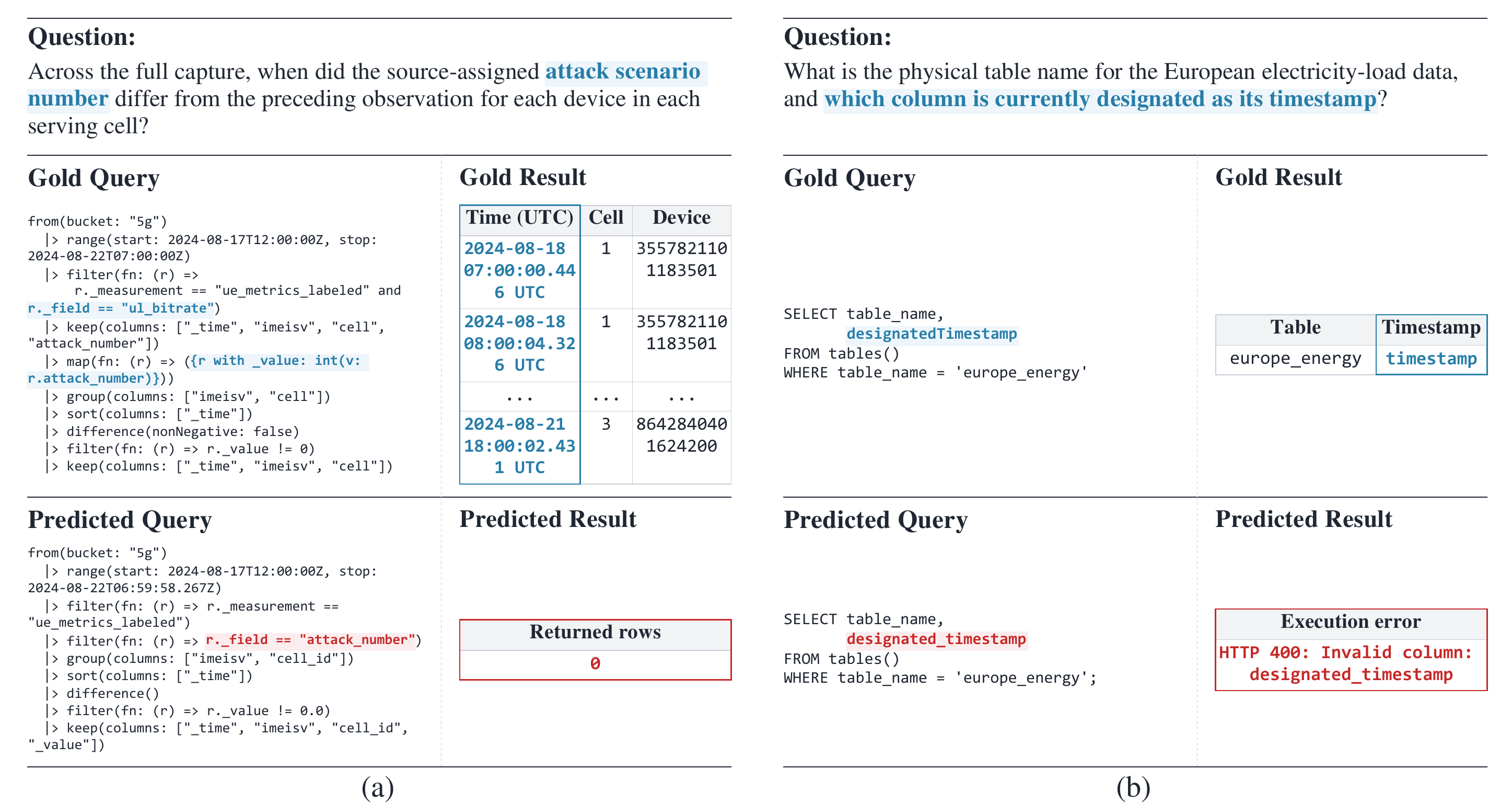}
    \caption{Two examples of schema linking errors. (a) The model treats the \texttt{attack\_number} tag as a field and filters on \texttt{\_field == "attack\_number"}, which removes all relevant records and produces an empty result. (b) The model uses \texttt{designated\_timestamp} instead of the metadata column \texttt{designatedTimestamp}, causing an invalid-column error and preventing retrieval of the designated timestamp column.}
    \label{fig:appendix-errors-schema-linking}
\end{figure}

\subsection{Details of the Ablation Study}
\label{appendix:ablation-study}
Here, we introduce the implementation details of the ablation study in Sec.~\ref{subsec:ablation-study}.
Although query syntax errors account for 75.29\% of all error cases, most of these cases involve two or more types of errors simultaneously. 
Therefore, to eliminate the interference from other error sources, we only select cases with syntax errors as the sole error type for the ablation study.
For QA pairs containing only syntax errors, we map the TSDBs to RDBs and organize the schemas following the same format as BIRD.
We then keep the original questions unchanged, manually annotate each question with a corresponding SQL gold query, and verify that the execution results are consistent with the original gold answers.
Based on this converted setting, we evaluate the model and observe that the execution accuracy on these error cases improves from 0\% to 32.21\%, which falls between the performance levels on BIRD and Spider 2.0-lite. 
This suggests that the poor performance of the model is primarily due to the model's inability to handle TSDB query syntax rather than the difficulty of the questions.

\begin{figure}[htbp]
    \centering
    \includegraphics[width=\linewidth]{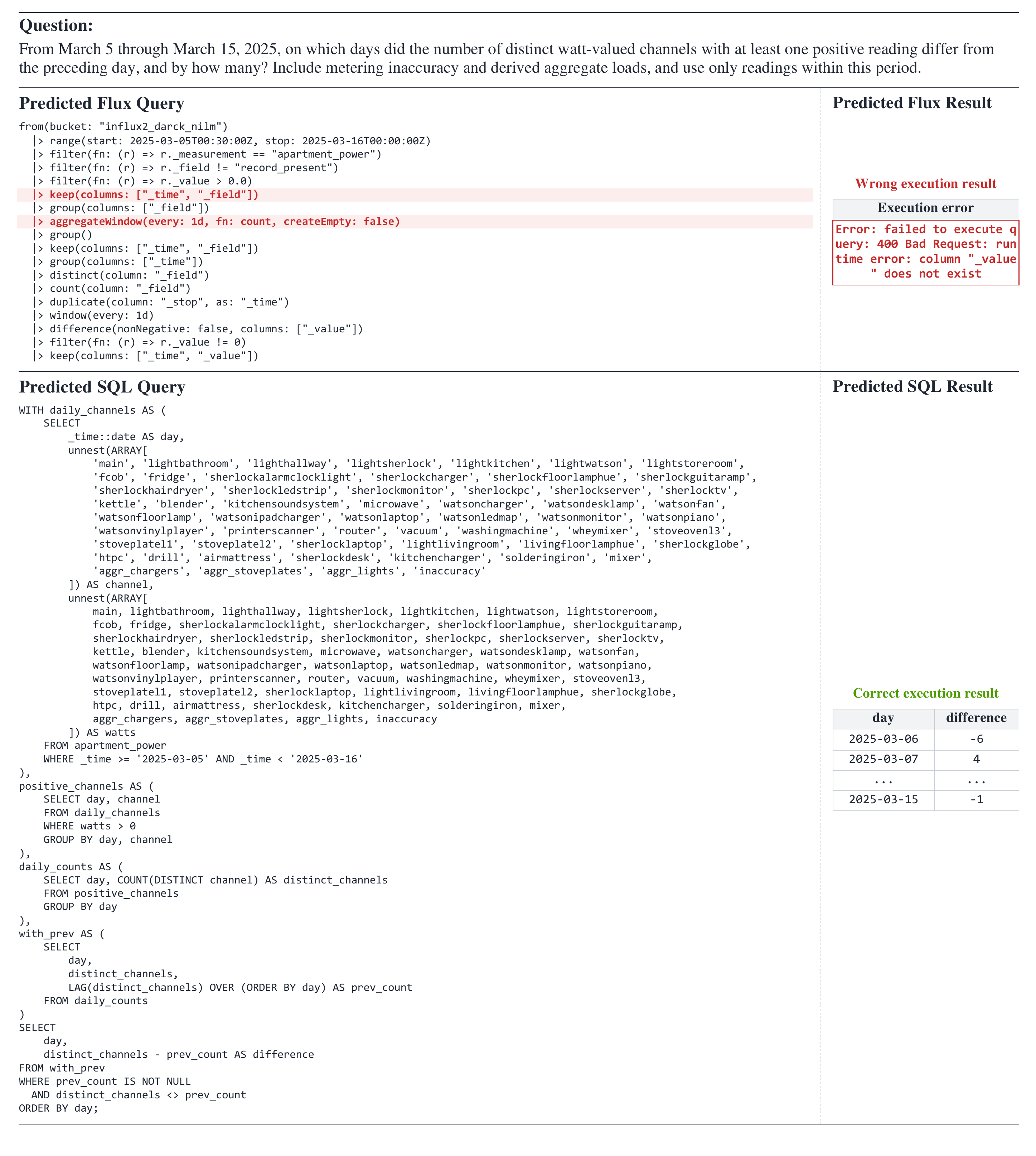}
    \caption{A case study on the differences in LLM capabilities between generating Flux for TSDBs and SQL for RDBs.
}
    \label{fig:syntax-conversion-example}
\end{figure}

To further demonstrate the difference in the model's proficiency between SQL and TSDB query syntax, we present a case study.
As shown in Fig.~\ref{fig:syntax-conversion-example}, the model-generated Flux query fails due to a syntax error, whereas the generated SQL query successfully produces the correct result, despite the SQL gold query, containing 766 tokens, requiring a much more complex implementation than its Flux counterpart, containing 255 tokens. 
This suggests that the model still lacks sufficient capability in understanding and generating TSDB query syntax, such as Flux.

\section{Prompts used in \bench~construction process}
\begin{promptbox}{The system prompt used for question generation in \bench~construction process}
    \textbf{[System]}\par

    \textbf{Task:}\par
    You are an assistant for constructing natural-language questions over time-series databases. 
    Using the query intents, target intent combination, database schema, and database context below, 
    generate one natural-language question (Q) for the target time-series database. 
    In this stage, generate Q only; do not generate a query statement (A) or query results.\par

    \textbf{Inputs:}\par

    \textbf{Query intents:}\par
    A question may involve one or more intents. I1--I4 are time-specific intents, 
    and I5--I9 are time-agnostic intents.\par

    \textbf{I1 -- Window aggregation and resampling:} 
    Resample or aggregate a time series over time windows or a new sampling grid, 
    yielding results for each window or grid.\par

    \textbf{I2 -- Temporal change analysis:} 
    Analyze temporal changes in values within a single time series.\par

    \textbf{I3 -- Time localization:} 
    Locate the time point(s) or interval at which an event or state occurs.\par

    \textbf{I4 -- Relationship analysis:} 
    Analyze relationships between two or more independently identifiable time series.\par

    \textbf{I5 -- Conditional filtering:} 
    Filter samples or events based on specified conditions.\par

    \textbf{I6 -- Statistical computation:} 
    Compute statistical values from samples or events.\par

    \textbf{I7 -- Sort \& rank:} 
    Rank candidates by one or more ordering criteria.\par

    \textbf{I8 -- Metadata query:} 
    Query metadata about time-series databases, such as schema information.\par

    \textbf{I9 -- Domain knowledge dependency:} 
    Require the use of specialized domain knowledge to answer the question.\par

    \textbf{Target intent combination for this generation:}\par
    \texttt{\{TARGET\_INTENT\_COMBINATION\}}\par

    \textbf{Database schema:}\par
    \texttt{\{DATABASE\_SCHEMA\}}\par

    \textbf{Database context:}\par
    \texttt{\{DATABASE\_CONTEXT\}}\par

    \textbf{Generation constraints:}\par
    1. The question must express a real, meaningful query need for the target database 
    and accurately reflect the specified target intent combination. Include an intent only 
    when it is genuinely required by the question.\par

    2. Use only entities, data meanings, and domain conventions supported by the schema 
    and database context, so that the question can be answered using the target database.\par

    3. Express the user's information need in natural, clear, and unambiguous English. 
    Do not mechanically turn physical field names or a query statement into a question, 
    and do not describe query implementation steps.\par

    4. Do not combine unrelated objectives merely to satisfy the target intent combination. 
    Do not assume schema elements, data meanings, domain knowledge, or database capabilities 
    that were not provided.\par

    5. Do not generate tasks such as forecasting, anomaly detection, or pattern clustering 
    that require separate algorithmic models and cannot be completed by querying the target TSDB alone.\par

    \textbf{Output format:}\par
    Return exactly one JSON object, with no explanation or Markdown code fence. 
    The \texttt{intents} array must list the intents actually used by Q and must match 
    the target intent combination:\par

    \textbf{[User]}\par
    Target intent combination: \texttt{\{I1, I5\}}\par
    Database schema: \texttt{\{schema\}}\par
    Database context: \texttt{\{context\}}\par

    \textbf{[Assistant]}\par
    \begin{lstlisting}[basicstyle=\ttfamily\small]
{
  "question": "What were the available-memory readings for farm140105, 
  farm140108, and farm140109 at each fifteen-minute checkpoint 
  during the final hour?",
  "intents": ["I1", "I5"]
}
\end{lstlisting}

    \textbf{[User]}\par
    Target intent combination: \texttt{\{TARGET\_INTENT\_COMBINATION\}}\par
    Database schema: \texttt{\{DATABASE\_SCHEMA\}}\par
    Database context: \texttt{\{DATABASE\_CONTEXT\}}
\end{promptbox}

\begin{promptbox}{The system prompt used for query generation in \bench~construction process}
    \textbf{[System]}\par

    \textbf{Task:}\par
    You are an assistant for constructing time-series database queries. 
    The natural-language question (Q) and its intent labels below have been finalized. 
    Using Q, the target database type, database schema, and database context, 
    generate a query statement (A) that correctly answers Q. 
    Do not change the meaning of Q.\par

    \textbf{Inputs:}\par

    \textbf{Finalized Q:}\par
    \texttt{\{REVIEWED\_QUESTION\}}\par

    \textbf{Confirmed intent labels:}\par
    \texttt{\{REVIEWED\_INTENTS\}}\par

    \textbf{Target database type:}\par
    \texttt{\{TARGET\_TSDB\}}\par

    \textbf{Database schema:}\par
    \texttt{\{DATABASE\_SCHEMA\}}\par

    \textbf{Database context:}\par
    \texttt{\{DATABASE\_CONTEXT\}}\par

    \textbf{Generation constraints:}\par

    1. Generate one query that can actually be executed in 
    \texttt{\{TARGET\_TSDB\}}. Use the query syntax supported by that database type; 
    do not default to the syntax of another database system. 
    The query must fully and accurately express the goal, conditions, and temporal 
    semantics of Q, and its result must answer Q.\par

    2. Use only objects that actually exist in the schema and data meanings and 
    domain conventions provided in the database context. Do not invent fields, data, 
    or query results.\par

    3. Do not add analysis goals, conditions, or result content that Q does not request, 
    and do not change Q to accommodate a particular query formulation.\par

    4. If the target database is accessible, execute A, check that execution succeeds, 
    that the result is non-empty and valid, and that the result answers Q; revise A 
    if the check fails. Unless Q explicitly asks for a nonexistent entity or an empty 
    result, an empty result does not count as a verified answer. If the database is 
    inaccessible, do not claim that A was executed or verified.\par

    5. If Q conflicts with the schema or database context, or if information needed 
    for a correct query is missing, do not guess a query statement.\par

    \textbf{Output format:}\par
    Return exactly one JSON object, with no explanation or Markdown code fence. 
    \texttt{question} must reproduce Q exactly, \texttt{intents} must match the 
    confirmed intent labels, and \texttt{answer} must contain a query for the 
    target database type. Record only evidence obtained from an actual execution 
    in \texttt{execution\_evidence}; if the query was not run, set 
    \texttt{executed} to \texttt{false} and do not invent results.\par

    \textbf{[User]}\par
    Finalized Q: \texttt{\{question\}}\par
    Confirmed intent labels: \texttt{\{I1, I5\}}\par
    Target database type: \texttt{\{TSDB\}}\par
    Database schema: \texttt{\{schema\}}\par
    Database context: \texttt{\{context\}}\par

    \textbf{[Assistant]}\par
    
    \begin{lstlisting}[basicstyle=\ttfamily\small]
{
  "target_tsdb": "Prometheus",
  "question": "Show the total memory allocation for the pods of 
  the admin user service on node k8s-node3 or k8s-node4 
  over the last hour.",
  "intents": ["I1", "I5"],
  "answer": "<executable query for the target database>",
  "execution_evidence": {
    "executed": false,
    "non_empty": null,
    "result_summary": null
  }
}
\end{lstlisting}

    \textbf{[User]}\par
    Finalized Q: \texttt{\{REVIEWED\_QUESTION\}}\par
    Confirmed intent labels: \texttt{\{REVIEWED\_INTENTS\}}\par
    Target database type: \texttt{\{TARGET\_TSDB\}}\par
    Database schema: \texttt{\{DATABASE\_SCHEMA\}}\par
    Database context: \texttt{\{DATABASE\_CONTEXT\}}
\end{promptbox}

\end{document}